\documentclass[sigconf]{acmart}
\usepackage{amsmath}
\usepackage{multirow}
\usepackage{tablefootnote}
\usepackage{graphicx}
\usepackage{subcaption}
\usepackage{enumitem}

\newcommand{\proposed}{\textsf{CMRL}}
\usepackage{balance}

\AtBeginDocument{%
  \providecommand\BibTeX{{%
    \normalfont B\kern-0.5em{\scshape i\kern-0.25em b}\kern-0.8em\TeX}}}

\copyrightyear{2023}
\acmYear{2023}
\setcopyright{acmlicensed}\acmConference[KDD '23]{Proceedings of the 29th ACM SIGKDD Conference on Knowledge Discovery and Data Mining}{August 6--10, 2023}{Long Beach, CA, USA}
\acmBooktitle{Proceedings of the 29th ACM SIGKDD Conference on Knowledge Discovery and Data Mining (KDD '23), August 6--10, 2023, Long Beach, CA, USA}
\acmPrice{15.00}
\acmDOI{10.1145/3580305.3599437}
\acmISBN{979-8-4007-0103-0/23/08}

\acmPrice{15.00}
\acmISBN{978-1-4503-XXXX-X/18/06}

\begin{document}

\title[Shift-Robust Molecular Relational Learning with Causal Substructure]{Shift-Robust Molecular Relational Learning \\ with Causal Substructure}

\author{Namkyeong Lee}
\affiliation{%
  \institution{KAIST}
  \city{Daejeon}
  \country{Republic of Korea}
}
\email{namkyeong96@kaist.ac.kr}

\author{Kanghoon Yoon}
\affiliation{%
  \institution{KAIST}
  \city{Daejeon}
  \country{Republic of Korea}
}
\email{ykhoon08@kaist.ac.kr}

\author{Gyoung S. Na}
\affiliation{%
  \institution{KRICT}
  \city{Daejeon}
  \country{Republic of Korea}
}
\email{ngs0@krict.re.kr}

\author{Sein Kim}
\affiliation{%
  \institution{KAIST}
  \city{Daejeon}
  \country{Republic of Korea}
}
\email{rlatpdlsgns@kaist.ac.kr}

\author{Chanyoung Park}
\affiliation{%
  \institution{KAIST}
  \city{Daejeon}
  \country{Republic of Korea}
}
\email{cy.park@kaist.ac.kr}
\authornote{
Corresponding author.
}

\renewcommand{\shortauthors}{Namkyeong Lee, Kanghoon Yoon, Gyoung S. Na, Sein Kim, \& Chanyoung Park}
\begin{abstract}
    Recently, molecular relational learning, whose goal is to predict the interaction behavior between molecular pairs, got a surge of interest in molecular sciences due to its wide range of applications.
    In this work, we propose \proposed~that is robust to the distributional shift in molecular relational learning by detecting the core substructure that is causally related to chemical reactions.
    To do so, we first assume a causal relationship based on the domain knowledge of molecular sciences and construct a structural causal model (SCM) that reveals the relationship between variables.
    Based on the SCM, we introduce a novel conditional intervention framework whose intervention is conditioned on the paired molecule.
    With the conditional intervention framework, our model successfully learns from the causal substructure and alleviates the confounding effect of shortcut substructures that are spuriously correlated to chemical reactions.
    Extensive experiments on various tasks with real-world and synthetic datasets demonstrate the superiority of \proposed~over state-of-the-art baseline models.
    Our code is available at \url{https://github.com/Namkyeong/CMRL}.
\end{abstract}


%

\maketitle

\section{Introduction}

\begin{figure}[t]
    \centering
    \includegraphics[width=0.9\columnwidth]{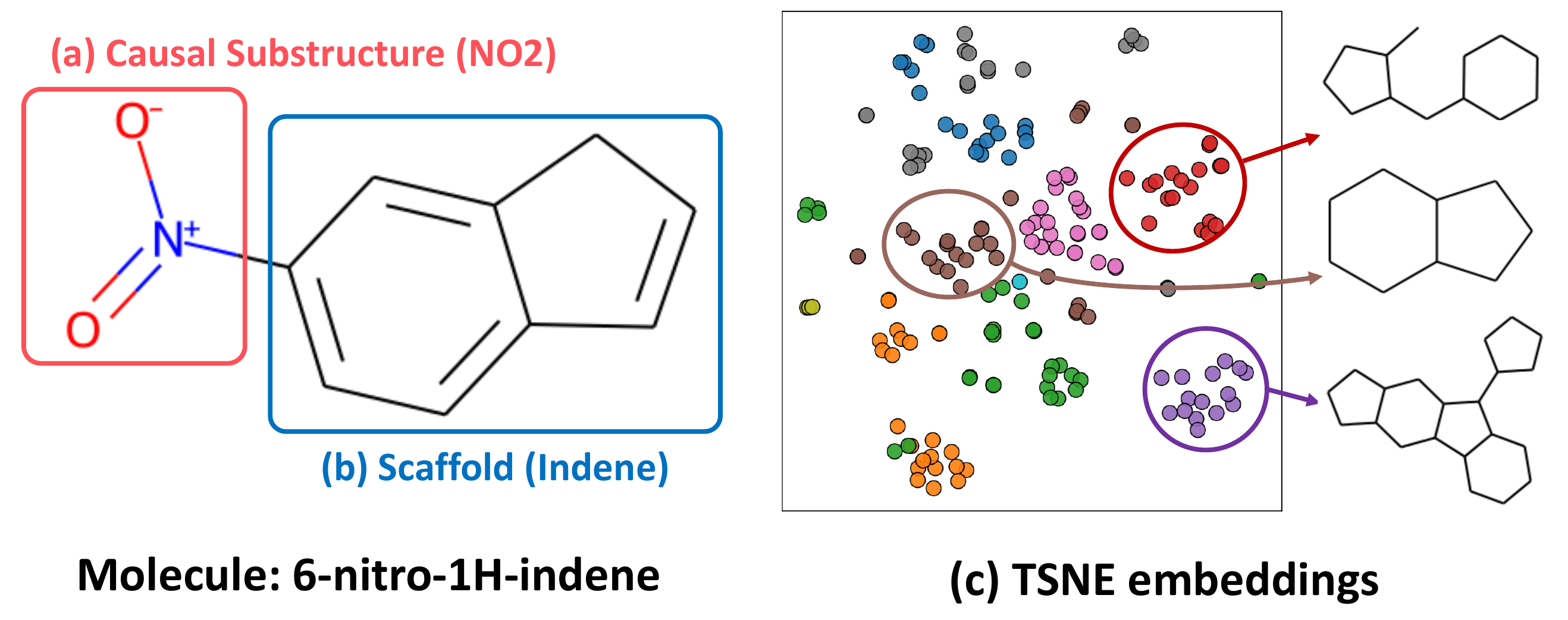} 
    \caption{(a) Nitrogen dioxide (NO2) is the causal substructure of 6-nitro-1H-indene for mutagenicity. (b) Indene is the scaffold of 6-nitro-1H-indene. (c) Visualization of TSNE embeddings of molecular fingerprints with various scaffolds.}
    \label{fig1}
    \vspace{-3ex}
\end{figure}

Along with recent advances in machine learning (ML) for scientific discovery, ML approaches have rapidly been adopted in molecular science, which encompasses the various field of biochemistry, molecular chemistry, medicine, and pharmacy \cite{mowbray2021machine,butler2018machine,sidey2019machine,vamathevan2019applications}.
Among the fields, molecular relational learning, which aims to predict the interaction behavior between molecule pairs, got a surge of interest among researchers due to its wide range of applications \cite{rozemberczki2021unified,rozemberczki2022chemicalx,rozemberczki2022moomin,pathak2021learning,lee2023conditional}.
For example, determining how medications will dissolve in various solvents (i.e., medication-solvent pair) and how different drug combinations will occur various synergistic/adverse effects (i.e., drug-drug pair) are crucial for discovering drugs in pharmaceutical sectors \cite{pathak2021learning,preuer2018deepsynergy,wang2021multi}.
Moreover, predicting the optical and photophysical properties of chromophores with various solvents (i.e., chromophore-solvent pair) is important for synthesizing new colorful materials in molecular chemistry \cite{joung2021deep}.

Despite the wide range of applications, traditional trial-error-based wet experiments require alternative computational approaches, i.e., ML, due to the following shortcomings:
1) Testing all possible combinations of molecule pairs requires expensive time/financial costs due to its combinatorial space \cite{preuer2018deepsynergy,rozemberczki2021unified},
2) Laboratory wet experiments are prone to human errors \cite{liu2020drugcombdb}, and
3) Certain types of tasks, e.g., predicting side effects of polypharmacy, require trials on humans that are highly risky \cite{nyamabo2021ssi}.
Among various ML-based approaches, graph neural networks (GNNs) have recently shown great success in molecular property prediction by representing a molecule as a graph, i.e., considering an atom and a bond in a molecule as a node and an edge in a graph, respectively \cite{gilmer2017neural,zhu2022unified,wang2021chemical,sun2021mocl,lee2022augmentation}.


However, as real-world graphs inherently contain redundant information and/or noise, it is challenging for GNNs to capture useful information for downstream tasks. Hence, recent studies focus on discovering core substructures that are highly related to the functionalities of the given graph~\cite{ying2019gnnexplainer,yuan2021explainability}.
Besides, it is well known that a molecule in particular explicitly contains the core substructure, i.e., functional group, that determines the chemical properties of a molecule regardless of other atoms that exist in a molecule, which facilitate a systematic prediction of chemical reactions and the behavior of chemical compounds \cite{jerry1992advanced,book2014compendium}.
For example, molecules with nitrogen dioxide (NO2) functional group (see Figure \ref{fig1} (a)) commonly exhibit the mutagenic property regardless of other irrelevant patterns such as carbon rings \cite{luo2020parameterized}.
In this work, we focus on the fact that functional groups also play a key role in molecular relational learning tasks. For example, the existence of hydroxyl groups, which are contained in molecules such as alcohol and glucose, increases the solubility of molecules in water (i.e., solvent) by increasing the polarity of molecules~\cite{delaney2004esol}, which gives a clue for the behavior of the interaction between a molecule-solvent pair.

The major challenge in learning from the core substructure of a molecule is the bias originated from the unpredictable nature of the data collection process \cite{sui2022causal,fan2022debiasing,fan2021generalizing}. 
That is, meaningless substructures (e.g., carbon rings) may be spuriously correlated with the label information (e.g., mutagenicity), incentivizing the model to capture such shortcuts for predicting the labels without discovering a causal substructure (e.g., nitrogen dioxide (NO2)).
However, as the shortcuts may change along with the shift in the data distribution, learning and relying on such shortcuts severely deteriorates the model's out-of-distribution (OOD) generalization ability \cite{wu2022discovering,li2022ood}, which is crucial in molecular sciences.
For example, discovering molecules that share common properties while belonging to totally different scaffold classes (see Figure \ref{fig1} (b)), i.e., scaffold hopping \cite{schneider1999scaffold,hu2017recent}, is helpful in developing novel chemotypes with improved properties \cite{zheng2021deep}, which is however non-trivial as different scaffolds have totally different distributions as shown in Figure \ref{fig1} (c).
Distribution shift is even more prevalent in molecular relational learning tasks, which is the main focus of this paper, whose training pairs are limited to the experimental data, while testing pairs are a combinatorially large universe of candidate molecules.

Recently, several methods have been proposed to learn from such biased graphs by discovering causal substructures based on structural causal model (SCM) \cite{pearl2000models} for various graph-level tasks such as rationale extraction \cite{wu2022discovering}, and graph classification \cite{fan2022debiasing,sui2022causal}.
Specifically, DIR \cite{wu2022discovering} generates a causal substructure by minimizing the variance between interventional risks thereby providing invariant explanations for the prediction of GNNs.
Moreover, CAL \cite{sui2022causal} learns to causally attend to core substructures for synthetic graph classification with various bias levels, while DisC \cite{fan2022debiasing} disentangles causal substructures from various biased color backgrounds.
However, how to discover causal substructures for molecular relational learning tasks has not yet been explored.

To this end, we first assume a causal relationship in molecular relational learning tasks based the domain knowledge in molecular sciences: \textit{Among the multiple substructures in a molecule, causal substructure on the model prediction varies depending on the paired molecule}.
That is, given two molecules, the causal substructure of a molecule is determined by not only the molecule itself, but also by its paired molecule.
We then construct an SCM that reveals the causal relationship between a pair of molecules (i.e., $\mathcal{G}^{1}$ and $\mathcal{G}^{2}$), the causal substructures of $\mathcal{G}^{1}$ (i.e., $\mathcal{C}^{1}$), and the shortcut substructures of $\mathcal{G}^{1}$ (i.e., $\mathcal{S}^{1}$) as shown in Figure \ref{fig2}.
Based on the SCM, we introduce a novel conditional intervention framework \cite{pearl2000models,pearl2014interpretation} for molecular relational learning tasks, whose intervention space on $\mathcal{C}^{1}$ is conditioned on the paired molecule $\mathcal{G}^{2}$.
By eliminating the confounding effect via the conditional intervention framework, 
we are able to estimate the true causal effect of $\mathcal{C}^{1}$ on the target variable $Y$ conditioned on the paired molecule $\mathcal{G}^{2}$.

Based on the analyzed causality, we propose a novel \textsf{C}ausal \textsf{M}olecular \textsf{R}elational \textsf{L}earner (\proposed), which predicts the interaction behavior between a pair of molecules $\mathcal{G}^{1}$ and $\mathcal{G}^{2}$ by maximizing the causal effects of the discovered causal feature $\mathcal{C}^{1}$ on the target variable $Y$. 
Specifically, \proposed~disentangles causal substructures $\mathcal{C}^{1}$ from shortcut substructures $\mathcal{S}^{1}$ conditioned on the paired molecule $\mathcal{G}^{2}$ in representation space by masking the shortcut features with noises.
Then, we adopt the conditional intervention framework by parametrizing the backdoor adjustment, which incentivizes the model to make a robust prediction with causal substructures $\mathcal{C}^{1}$, paired molecules $\mathcal{G}^{2}$, and various shortcut substructures $\mathcal{S}^{1}$ generated conditionally on the paired molecule $\mathcal{G}^{2}$.
By doing so, \proposed~learns the true causality between the causal feature and the model prediction in various molecular relational learning tasks, regardless of various shortcut features and distribution shifts.

Our extensive experiments on fourteen real-world datasets in various tasks, i.e., molecular interaction prediction, drug-drug interaction, and graph similarity learning, demonstrate the superiority of \proposed~not only for molecular relational learning tasks but also the general graph relational learning tasks.
Moreover, ablation studies verify that \proposed~successfully adopts causal inference for molecular relational learning with the novel conditional intervention framework.
The further appeal of \proposed~is explainability, whose explanation is provided by the causal substructure invoking the chemical reaction, which can further accelerate the discovery process in molecular science.
To the best of our knowledge, \proposed~is the first work that adopts causal inference for molecular relational learning tasks.

\vspace{-1.5ex}
\section{Related Work}
\subsection{Molecular Relational Learning}
In molecular sciences, predicting the reaction behaviors of various chemicals is important due to its wide range of applications.
In this paper, we focus on the \textit{molecular pairs}, i.e., predicting the interaction behavior between molecular pairs in a specific context of interest, including molecular interaction prediction and drug-drug interaction (DDI) prediction.

\noindent \textbf{Molecular Interaction Prediction.}
In molecular interaction prediction tasks, the model aims to learn the properties of molecules produced by chemical reactions or the properties of the reaction itself, which is crucial in discovering/designing new molecules.
For example, predicting the solubility of molecules in various solvents is directly related to drug discovery, while the optical/photophysical properties of molecules in various solvents are important in designing colorful materials synthesis.
Recently, Delfos \cite{lim2019delfos} proposes to predict the solvation free energy, which determines the solubility of molecules, with recurrent neural networks and attention mechanism with SMILES string \cite{weininger1988smiles} as input.
Inspired by the recent success of GNNs in molecular property prediction, CIGIN \cite{pathak2020chemically} learns to predict the solvation free energy by modeling the molecules as the graph structure.
CIGIN utilizes a co-attention map, which indicates the pairwise importance of atom interaction, enhancing the interpretability of chemical reactions.
\citet{joung2021deep} proposes to predict various optical/photophysical properties of chromophores with the representations obtained from graph convolutional networks (GCNs) \cite{kipf2016semi}.

\noindent \textbf{Drug-Drug Interaction Prediction.}
In DDI prediction tasks, the model aims to learn whether two combinations of drugs will occur side effects, which is a common interest in pharmaceutical sectors.
Traditional ML methods predict DDI by comparing Tanimoto similarity \cite{bajusz2015tanimoto} of drug fingerprints \cite{vilar2012drug} or exhibiting similarity-based features \cite{kastrin2018predicting}, under the assumption that similar drugs are likely to interact with each other.
Recently, MHCADDI \cite{deac2019drug} proposes a graph co-attention mechanism, which aggregates the messages from not only the atoms in the drug but also from the atoms in the paired drug.
On the other hand, SSI-DDI \cite{nyamabo2021ssi} identifies the interaction between drugs with co-attention between substructures in each drug.
MIRACLE \cite{wang2021multi} casts the tasks as link prediction tasks between drugs, by constructing graphs based on interaction data. Therefore, MIRACLE leverages a multi-view graph, in which one reveals the interaction history and the other is the drug structure.

Despite the success of existing studies in various molecular relational learning tasks,
they do not consider the causal relationship between the causal substructure and the target predictions,
which may fail in real-world scenarios, e.g., OOD.
Moreover, previous methods tend to target a single specific task, raising doubts about the generalization capability of the methods.
Therefore, we aim to build a model for various molecular relational learning tasks by discovering causal substructures with causal relationships.

\vspace{-1.5ex}
\subsection{Causal Inference}
Causal inference is an effective tool for learning the causal relationship between various variables \cite{glymour2016causal,pearl2018book}.
Recently, several researchers have introduced causal inference in machine learning \cite{bengio2019meta}, successfully addressing the challenges in various domains such as computer vision \cite{gong2016domain,zhang2020causal,yue2020interventional,magliacane2018domain}, natural language processing \cite{joshi2022all,wang2021identifying,eisenstein2022informativeness}, and reinforcement learning \cite{bareinboim2015bandits,gasse2021causal}.
Inspired by the success of causal inference in various domains, it has been adopted in the graph domain, showing robustness in biased datasets and distribution shifts \cite{fan2022debiasing,sui2022causal,wu2022discovering,li2022learning}.
Specifically, DIR \cite{wu2022discovering} provides interpretations on GNNs' predictions by discovering invariant rationales in various interventional distributions.
For graph classification tasks, GIL \cite{li2022learning} proposes to capture the invariant relationship between causal substructure and labels in various environments, thereby achieving the capability of OOD generalization under distribution shifts.
Moreover, DisC \cite{fan2022debiasing} and CAL \cite{sui2022causal} improve the model robustness by making predictions with disentangled causal substructure in the graph and various shortcut substructures from other graphs for superpixel graph classification and synthetic graph classification, respectively.
Despite the successful adaptation of causal inference in the graph domain, previous studies have focused on tasks that require only a single graph, hanging a question mark over the effectiveness of causal inference in molecular relational learning tasks.
To the best of our knowledge, \proposed~is the first work that adopts the causal inference into molecular relational learning tasks.

\section{Problem Statement}
\noindent \textbf{Notations.}
Let $\mathcal{G} = (\mathcal{V}, \mathcal{E})$ denote a molecular graph, where $\mathcal{V} = \{v_{1},\ldots, v_{N}\}$ represents the set of atoms, and $\mathcal{E} \subseteq \mathcal{V} \times \mathcal{V}$ represents the set of bonds. $\mathcal{G}$ is associated with a feature matrix $\mathbf{X} \in \mathbb{R}^{N \times F}$, and an adjacency matrix $\mathbf{A} \in \mathbb{R}^{N \times N}$ where $\mathbf{A}_{ij} = 1$ if and only if $(v_i, v_j) \in \mathcal{E}$ and $\mathbf{A}_{ij} = 0$ otherwise.
We denote the atom representation matrix as $\mathbf{H} \in \mathbb{R}^{N \times 2d}$ whose $i$-th row, i.e., $\mathbf{H}_{i} = \mathbf{H}\left[i:\right]$, indicates the representation of atom $i$.

\smallskip
\noindent \textbf{Task: Molecular relational learning.}
Given a set of molecular pairs $\mathcal{D} = \{(\mathcal{G}_1^{1}, \mathcal{G}_1^{2}), (\mathcal{G}_2^{1}, \mathcal{G}_2^{2}), \ldots, (\mathcal{G}_n^{1}, \mathcal{G}_n^{2}) \}$ and the corresponding target values $\mathbb{Y} = \{\mathbf{Y}_1, \mathbf{Y}_2, \ldots, \mathbf{Y}_n \}$, 
we aim to train a model $\mathcal{M}$ that predicts the target values of a given arbitrary molecular pair, i.e., $\mathbf{Y}_{i} = \mathcal{M}(\mathcal{G}_{i}^{1}, \mathcal{G}_{i}^{2})$.
The target $\mathbf{Y}$ is a scalar value, i.e., $\mathbf{Y} \in (-\infty, \infty)$, for regression tasks, while it is a binary class label, i.e., $\mathbf{Y} \in \{0, 1\}$, for classification tasks.

\section{Methodology}

\subsection{A Causality in Molecular relational learning}

\begin{figure}[t]
    \centering
    \includegraphics[width=0.9\columnwidth]{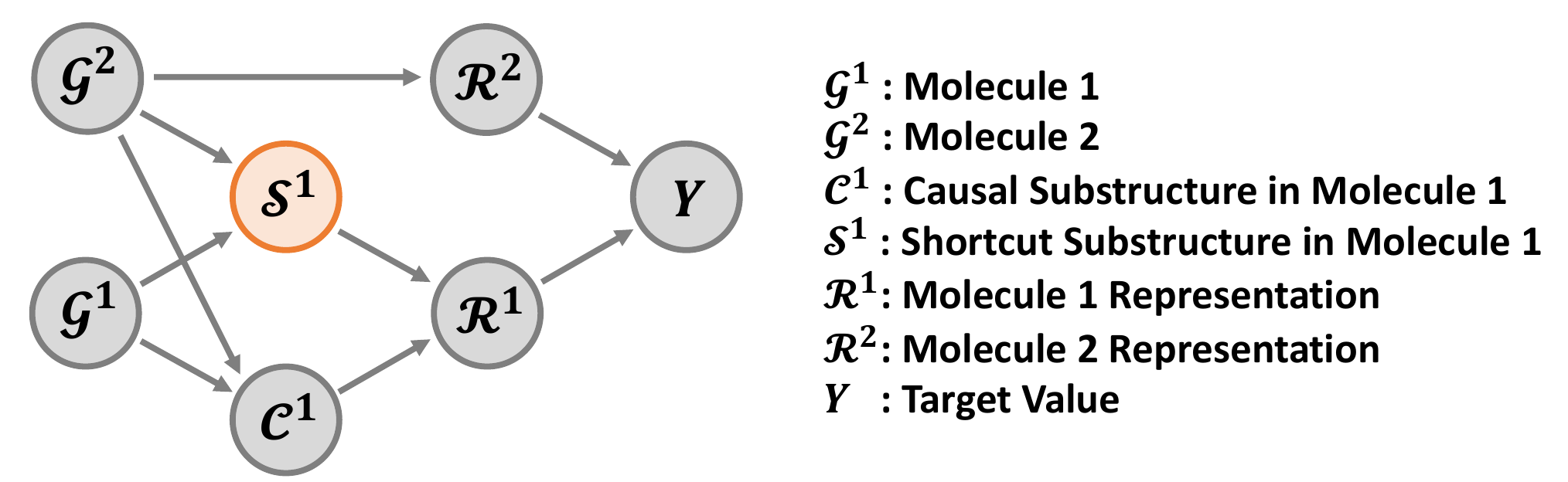} 
    \caption{Structural causal model for molecular relational learning.}
    \label{fig2}
\end{figure}

We formulate causalities in the decision-making process of GNNs for molecular relational learning tasks and construct a structural causal model (SCM) \cite{pearl2000models} in Figure \ref{fig2}, which reveals the causal relationship between seven variables: a molecule $\mathcal{G}^{1}$, another molecule $\mathcal{G}^{2}$, a causal substructure $\mathcal{C}^{1}$ of molecule $\mathcal{G}^{1}$, a shortcut substructure $\mathcal{S}^{1}$ of molecule $\mathcal{G}^{1}$, representation $\mathcal{R}^{1}$ of molecule $\mathcal{G}^{1}$, representation $\mathcal{R}^{2}$ of molecule $\mathcal{G}^{2}$, and the target value $\mathbf{Y}$.
Each link, i.e., $\rightarrow$, in SCM indicates a causal-effect relationship, i.e., \textsf{cause} $\rightarrow$ \textsf{effect}.
We give the following explanations for each causal-effect relationship:

\begin{itemize}[leftmargin=.15in]
    \item \underline{$\mathcal{G}^{1} \boldsymbol{\rightarrow} \mathcal{C}^{1} \boldsymbol{\leftarrow} \mathcal{G}^{2}$}:
$\mathcal{C}^{1}$ is a causal substructure in molecule $\mathcal{G}^{1}$ that determines the model decision of when to interact with molecule $\mathcal{G}^{2}$. This causal relationship is established since the causal substructure $\mathcal{C}^{1}$ is determined by not only the molecule $\mathcal{G}^{1}$ itself, but also by the paired molecule $\mathcal{G}^{2}$, which is well-known domain knowledge in molecular sciences.
For example, when determining the solubility of molecules in various solvents (i.e., paired molecule), C-CF3 substructure plays a key role when the solvent is water \cite{purser2008fluorine}, while it may not be the case when the solvent is oil.

\item \underline{$\mathcal{G}^{1} \boldsymbol{\rightarrow} \mathcal{S}^{1} \boldsymbol{\leftarrow} \mathcal{G}^{2}$}:
$\mathcal{S}^{1}$ is a shortcut substructure in molecule $\mathcal{G}^{1}$ that is spuriously correlated with label information $Y$. Since the causal substructure $\mathcal{C}^{1}$ varies regarding the paired molecule $\mathcal{G}^{2}$, the shortcut substructure $\mathcal{S}^{1}$ also varies considering the paired molecule $\mathcal{G}^{2}$.

\item \underline{$\mathcal{C}^{1} \boldsymbol{\rightarrow} \mathcal{R}^{1} \boldsymbol{\leftarrow} \mathcal{S}^{1}$}:
The variable $\mathcal{R}^{1}$ is the representation of molecule $\mathcal{G}^{1}$, which is conventionally obtained by incorporating both causal substructure $\mathcal{C}^{1}$ and shortcut substructure $\mathcal{S}^{1}$ as inputs.

\item \underline{$\mathcal{G}^{2} \boldsymbol{\rightarrow} \mathcal{R}^{2}$}:
The variable $\mathcal{R}^{2}$ is the representation of molecule $\mathcal{G}^{2}$, which is obtained by encoding molecule $\mathcal{G}^{2}$.

\item \underline{$\mathcal{R}^{1} \boldsymbol{\rightarrow} \mathbf{Y} \boldsymbol{\leftarrow} \mathcal{R}^{2}$}:
In molecular relational learning tasks, the model prediction is determined by not only the molecule $\mathcal{G}^{1}$ but also the paired molecule $\mathcal{G}^{2}$.
For example, when predicting the solubility of molecules in various solvents, polar solutes tend to have high solubility in polar solvents, while they are more likely to have low solubility in non-polar solvents.
Therefore, the target variable $\mathbf{Y}$ is affected by two factors $\mathcal{R}^{1}$ and $\mathcal{R}^{2}$.

\end{itemize}

\smallskip
Based on our assumed SCM, we find out that there exist four backdoor paths that confound the model to learn from true causalities between $\mathcal{C}^{1}$ and $\mathbf{Y}$, i.e., $\mathcal{C}^{1} \leftarrow \mathcal{G}^{1} \rightarrow \mathcal{S}^{1} \leftarrow \mathcal{G}^{2} \rightarrow \mathcal{R}^{2} \rightarrow \mathbf{Y}$, $\mathcal{C}^{1} \leftarrow \mathcal{G}^{2} \rightarrow \mathcal{R}^{2} \rightarrow \mathbf{Y}$, $\mathcal{C}^{1} \leftarrow \mathcal{G}^{2} \rightarrow \mathcal{S}^{1} \rightarrow \mathcal{R}^{1} \rightarrow \mathbf{Y}$, and $\mathcal{C}^{1} \leftarrow \mathcal{G}^{1} \rightarrow \mathcal{S}^{1} \rightarrow \mathcal{R}^{1} \rightarrow \mathbf{Y}$, leaving $\mathcal{G}^{2}$ and $\mathcal{S}^{1}$ as the backdoor criteria.
{
However, thanks to the nature of molecular relational learning tasks, i.e., $\mathcal{G}^{2}$ is given and utilized during model prediction, all the backdoor paths except for $\mathcal{C}^{1} \leftarrow \mathcal{G}^{1} \rightarrow \mathcal{S}^{1} \rightarrow \mathcal{R}^{1} \rightarrow \mathbf{Y}$ are blocked by conditioning on $\mathcal{G}^{2}$.
}
Therefore, we should now eliminate the confounding effect of $\mathcal{S}^{1}$, which is the only remaining element for backdoor criteria, on the model prediction and make the model utilize the causal substructure $\mathcal{C}^{1}$ and the paired molecule $\mathcal{G}^{2}$.

\subsection{Backdoor Adjustment}
\label{sec: Backdoor Adjustment}
One simple way to alleviate the confounding effect of $\mathcal{S}^{1}$ is to collect molecules with a causal substructure $\mathcal{C}^{1}$ and various shortcut substructures $\tilde{\mathcal{S}}^{1}$, which is impossible due to its expensive time/financial costs \cite{zhang2020causal}.
Fortunately, the causal theory provides an efficient tool, i.e., backdoor adjustment, for the model to learn intervened distribution $\tilde{P}(\mathbf{Y} | \mathcal{C}^{1}, \mathcal{G}^{2}) = P(\mathbf{Y}|do(\mathcal{C}^{1}), \mathcal{G}^{2})$ by eliminating backdoor paths, instead of directly learning confounded distribution $P(\mathbf{Y}|\mathcal{C}^{1}, \mathcal{G}^{2})$.
Specifically, the backdoor adjustment alleviates the confounding effect of $\mathcal{S}^{1}$ by making a model prediction with ``intervened molecules,'' which is created by stratifying the confounder into pieces $\mathcal{S}^{1} = \{ s \}$ and pairing the causal substructure $\mathcal{C}^{1}$ with every stratified confounder pieces $\{s\}$.
More formally, the backdoor adjustment can be formulated as follows:

\begin{equation}
\small
\begin{split}
    P(\mathbf{Y}|do(\mathcal{C}^{1}), \mathcal{G}^{2}) &= \tilde{P}(\mathbf{Y}|\mathcal{C}^{1}, \mathcal{G}^{2})\\
    &= \sum_{s}{\tilde{P}(\mathbf{Y}|\mathcal{C}^{1}, \mathcal{G}^{2}, s) \cdot \tilde{P}(s|\mathcal{C}^{1}, \mathcal{G}^{2})} \,\, \text{(Bayes' Rule)}\\
    &= \sum_{s}{\tilde{P}(\mathbf{Y}|\mathcal{C}^{1}, \mathcal{G}^{2}, s) \cdot \tilde{P}(s|\mathcal{G}^{2})} \,\, \text{(Independence)}\\
    &= \sum_{s}{{P}(\mathbf{Y}|\mathcal{C}^{1}, \mathcal{G}^{2}, s) \cdot {P}(s|\mathcal{G}^{2})}, \\
\end{split}
\label{Eq: Backdoor adjustment}
\end{equation}

\noindent where $P(\mathbf{Y}|\mathcal{C}^{1}, \mathcal{G}^{2}, s)$ is the conditional probability given causal substructure $\mathcal{C}^{1}$, paired molecule $\mathcal{G}^{2}$, and confounder $s$, while $P(s|\mathcal{G}^{2})$ indicates the conditional probability of confounder $s$ given the paired molecule $\mathcal{G}^{2}$.
With the formalized intervened distribution, we observe that the model should make a prediction with causal substructure $\mathcal{C}^{1}$ in molecule $\mathcal{G}^{1}$ and the paired molecule $\mathcal{G}^{2}$ incorporating all possible shortcut substructures $s$, i.e., $P(\mathbf{Y}|\mathcal{C}^{1}, \mathcal{G}^{2}, s)$.
By doing so, the backdoor adjustment enables the model to make robust and invariant predictions by learning from various stratified shortcut substructures $s$.
On the other hand, different from previous works \cite{sui2022causal,fan2022debiasing,wu2022discovering}, we narrow down the intervention space by conditioning confounders $s$ on the paired molecule $\mathcal{G}^{2}$, i.e., $P(s|\mathcal{G}^{2})$, which is supposed to be wider than necessary due to its combinatorial nature.
Considering that the model prediction is made by not only $\mathcal{G}^{1}$ but also by $\mathcal{G}^{2}$, constraining the intervention space prevents the model from suffering from the involvement of unnecessary information, which may disrupt model training.
We argue that a conditional intervention framework is a key to the success of \proposed, whose importance will be demonstrated in Section \ref{sec: Ablation Studies}.


\subsection{Model Architecture}
\label{sec: Model Architecture}

\begin{figure}[t]
    \centering
    \includegraphics[width=0.8\columnwidth]{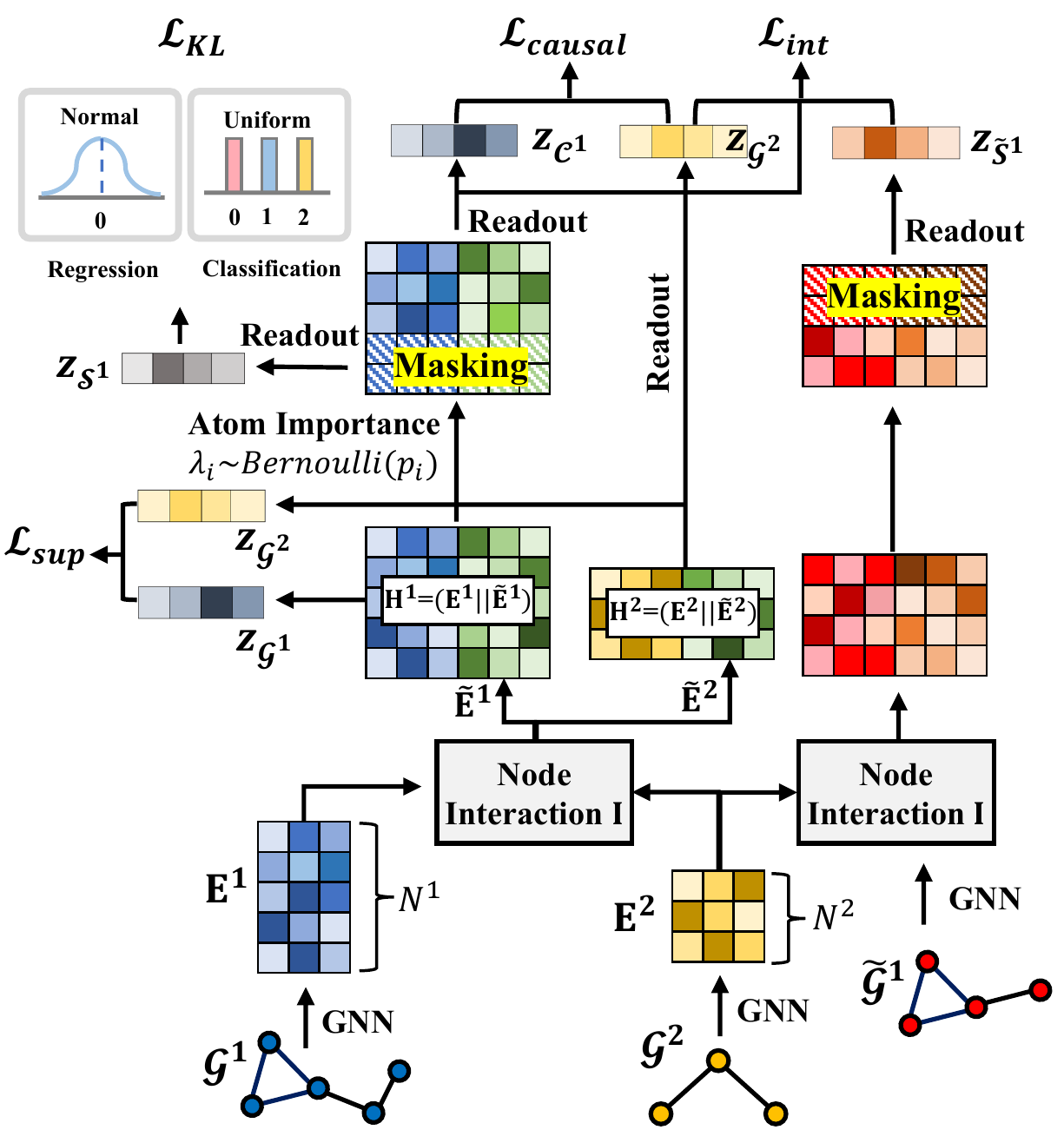} 
    \caption{Overall model architecture.}
    \label{fig:model architecture}
    \vspace{-2ex}
\end{figure}

In this section, we implement our framework \proposed~ based on the architecture of CIGIN \cite{pathak2020chemically}, which is a simple and intuitive architecture designed for molecular relational learning tasks.
In a nutshell, CIGIN aims to generate the representations $\mathcal{R}^{1}$ and $\mathcal{R}^{2}$ of given molecular pairs $\mathcal{G}^{1}$ and $\mathcal{G}^{2}$ with an interaction map, which indicates the importance of pairwise atom-level interactions between $\mathcal{G}^{1}$ and $\mathcal{G}^{2}$.
Specifically, given a pair of molecules $\mathcal{G}^{1} = (\mathbf{X}^{1}, \mathbf{A}^{1})$ and $\mathcal{G}^{2} = (\mathbf{X}^{2}, \mathbf{A}^{2})$, we first obtain the atom representation matrix of each molecule as follows:
\begin{equation}
\small
    \mathbf{E}^{1} = \text{GNN}(\mathbf{X}^{1}, \mathbf{A}^{1}), \,\,\,\,
    \mathbf{E}^{2} = \text{GNN}(\mathbf{X}^{2}, \mathbf{A}^{2})
\end{equation}
where $\mathbf{E}^{1} \in \mathbb{R}^{N^1\times d}$ and $\mathbf{E}^{2} \in \mathbb{R}^{N^2\times d}$ are atom representation matrices of molecules $\mathcal{G}^{1}$ and $\mathcal{G}^{2}$, respectively, and $N^1$ and $N^2$ denote the number of atoms in molecule $\mathcal{G}^{1}$ and $\mathcal{G}^{2}$, respectively.
Then, the interaction between two molecules $\mathcal{G}^{1}$ and $\mathcal{G}^{2}$ is modeled via an interaction map $\mathbf{I} \in \mathbb{R}^{N^1\times N^2}$, which is defined as follows: $\mathbf{I}_{ij} = \mathsf{sim}(\mathbf{E}^{1}_{i}, \mathbf{E}^{2}_{j})$, where $\mathsf{sim}(\cdot, \cdot)$ indicates the cosine similarity.
Given the interaction map $\mathbf{I}$, we calculate another atom representation matrix 
$\tilde{\mathbf{E}}^{1} \in \mathbb{R}^{N^1\times d}$ and $\tilde{\mathbf{E}}^{2} \in \mathbb{R}^{N^2\times d}$ for each molecule $\mathcal{G}^{1}$ and $\mathcal{G}^{2}$ as follows:
\begin{equation}
    \tilde{\mathbf{E}}^{1}  = \mathbf{I} \cdot \mathbf{E}^{2}, \,\,\,\,
    \tilde{\mathbf{E}}^{2} = \mathbf{I}^\top \cdot \mathbf{E}^{1}
\end{equation}
where $\cdot$ indicates matrix multiplication between two matrices.
Thus, $\tilde{\mathbf{E}}^{1}$ is an atom representation matrix for molecule $\mathcal{G}^{1}$, which also contains the information about the paired molecule $\mathcal{G}^{2}$, and likewise for $\tilde{\mathbf{E}}^{2}$.
Then, the atom representation matrix $\mathbf{H}^{1} \in \mathbb{R}^{N^1\times 2d}$ for molecule $\mathcal{G}^{1}$ is generated by concatenating two representation matrices $\tilde{\mathbf{E}}^{1}$ and $\mathbf{E}^{1}$, i.e., $\mathbf{H}^{1} = (\mathbf{E}^{1} || \tilde{\mathbf{E}}^{1})$.
We also generate the atom representation matrix $\mathbf{H}^{2} \in \mathbb{R}^{N^2\times 2d}$ for molecule $\mathcal{G}^{2}$ in a similar way.
Finally, we obtain molecular level representations $z_{\mathcal{G}^{1}}$ and $z_{\mathcal{G}^{2}}$ for molecules $\mathcal{G}^{1}$ and $\mathcal{G}^{1}$, respectively, by pooling each representation matrix with the Set2Set readout function \cite{vinyals2015order}.
The overall model architecture is depicted in Figure \ref{fig:model architecture}.

\subsection{Causal Molecular Relational Learner}
In this section, we propose \textsf{C}ausal \textsf{M}olecular \textsf{R}elational \textsf{L}earner (\proposed) framework to implement the backdoor adjustment proposed in Section \ref{sec: Backdoor Adjustment} with the architecture described in Section \ref{sec: Model Architecture}.

\subsubsection{Disetangling with Atom Representation Masks.}
To successfully adopt a causal intervention framework in the molecular relational learning task, it is crucial to separate the causal substructure $\mathcal{C}^{1}$ and the shortcut substructure $\mathcal{S}^{1}$ from $\mathcal{G}^{1}$.
However, since the molecule is formed with various domain knowledge such as valency rules, it is not trivial to explicitly manipulate the graph structure \cite{sui2022causal}.
Moreover, as depicted in Figure \ref{fig2}, causal substructure $\mathcal{C}^{1}$ is determined by not only the molecule $\mathcal{G}^{1}$ itself, but also the paired molecule $\mathcal{G}^{2}$.
To this end, we propose a model to separate the causal substructure $\mathcal{C}^{1}$ and the shortcut substructure $\mathcal{S}^{1}$ from $\mathcal{G}^{1}$ in representation space, by masking the atom representation matrix $\mathbf{H}^{1}$ that contains information regarding both molecules $\mathcal{G}^{1}$ and $\mathcal{G}^{2}$.
Specifically, given an atom $i$'s embedding $\mathbf{H}^{1}_{i}$, we parametrize the importance $p_i$ of atom $i$ in model prediction with MLP as follows:
\begin{equation}
\small
    p_i = \text{MLP}(\mathbf{H}^{1}_{i}).
\end{equation}
With the calculated importance $p_i$ of atom $i$, we obtain the atom $i$'s representation in causal substructure $\mathbf{C}^{1}_{i}$ and shortcut substructure $\mathbf{S}^{1}_{i}$ with a mask $\epsilon$ that is sampled from the normal distribution as follows:
\begin{equation}
\small
    \mathbf{C}^{1}_{i} = \lambda_{i}\mathbf{H}^{1}_{i} + (1 - \lambda_{i}) \epsilon, \,\,\,\,
    \mathbf{S}^{1}_{i} = (1 - \lambda_{i}) \mathbf{H}^{1}_{i},
\end{equation}
where $\lambda_{i} \sim \text{Bernoulli}(p_i)$ and $\epsilon \sim N(\mu_{\mathbf{H}^{1}}, \sigma_{\mathbf{H}^{1}}^{2})$.
Note that $\mu_{\mathbf{H}^{1}}$ and $\sigma_{\mathbf{H}^{1}}^{2}$ indicate the mean and variance of atom embeddings in $\mathbf{H}^{1}$, respectively.
By doing so, we expect the model to automatically learn a causal substructure $\mathcal{C}^{1}$ that is causally related to the target variable $\mathbf{Y}$, while masking the shortcut substructure $\mathcal{S}^1$ that confounds the model prediction.
Moreover, since sampling $\lambda_{i}$ from Bernoulli distribution is a non-differentiable operation, we adopt the Gumbel sigmoid \cite{jang2016categorical,maddison2016concrete} approach for sampling $\lambda_{i}$ as follows:
\begin{equation}
    \lambda_{i} = \text{Sigmoid}(1/t\log[{p_i/(1-p_i)}]+\log{[u/(1-u)]}),
\end{equation}
where $u \sim \text{Uniform}(0, 1)$, and $t$ is the temperature hyperparameter.
With the learned causal representation matrix $\mathbf{C}^{1}$ and shortcut representation matrix $\mathbf{S}^{1}$, we obtain the substructure level representation $z_{\mathcal{C}^{1}}$ and $z_{\mathcal{S}^{1}}$ with the Set2Set readout function.

Now, we present the objective function for learning causal representation matrix $\mathbf{C}^{1}$ and shortcut representation matrix $\mathbf{S}^{1}$ to capture the causal and shortcut features in the molecule $\mathcal{G}^{1}$, respectively.
For causal representation matrix $\mathbf{C}^{1}$, we aim to discover causal substructure $\mathcal{C}^{1}$ that is causally related to the model prediction.
Therefore, we expect the model to make a similar/same prediction given a pair of causal substructure $\mathcal{C}^{1}$ and molecule $\mathcal{G}^{2}$ with the prediction given a pair of molecules $\mathcal{G}^{1}$ and $\mathcal{G}^{2}$.
To this end, we propose a prediction loss $\mathcal{L}_{causal}(\mathbf{Y}, z_{\mathcal{C}^{1}}, z_{\mathcal{G}^{2}})$, which can be modeled as the cross entropy loss for classification and the root mean squared error loss for regression.
On the other hand, we expect the shortcut substructure $\mathcal{S}^{1}$ to have no information regarding predicting the label $\mathbf{Y}$.
Therefore, we propose the shortcut substructure to mimic non-informative distribution, which provides no information on target $\mathbf{Y}$ at all.
Specifically, we propose to minimize KL divergence loss $\mathcal{L}_{KL}(\mathbf{Y}_{rand}, z_{\mathcal{S}^{1}})$, where the label $\mathbf{Y}_{rand}$ is given as the uniform distribution for classification and standard normal distribution for regression.
By optimizing the proposed loss terms, i.e., $\mathcal{L}_{causal}$ and $\mathcal{L}_{KL}$, \proposed~effectively disentangles the causal substructure $\mathcal{C}^{1}$ and the shortcut substructure $\mathcal{S}^{1}$ from the molecule $\mathcal{G}^{1}$.

\subsubsection{Conditional Causal Intervention.}
\label{sec: Conditional Causal Intervention}
In this section, we implement the backdoor adjustment described in Section \ref{sec: Backdoor Adjustment}, which successfully alleviates the confounding effect of the shortcut substructure $\mathcal{S}^{1}$.
One straightforward approach for the backdoor adjustment is to generate an intervened molecule structure, i.e., molecular generation with causal substructure $\mathcal{C}^{1}$ and various shortcut substructures $\tilde{\mathcal{S}}^{1}$.
However, directly applying such an intuitive approach is not trivial since 1) the molecules exist on the basis of various domain knowledge in molecular sciences, and 2) our intervention space on $\mathcal{C}^{1}$ should be conditioned on the paired molecule $\mathcal{G}^{2}$ as shown in Equation \ref{Eq: Backdoor adjustment}.
To this end, we propose a simple yet effective approach for the conditional causal intervention, whose intervention is done implicitly in the representation space conditioned on molecule $\mathcal{G}^{2}$.
Specifically, given molecule $\mathcal{G}^{2}$, we obtain shortcut substructure $\tilde{\mathcal{S}}^{1}$ by modeling the interaction between other molecules $\tilde{\mathcal{G}}^{1}$ in the training data $\mathcal{D}$ and molecule $\mathcal{G}^{2}$.
After obtaining the representation of shortcut substructures $z_{\tilde{\mathcal{S}}^{1}}$, we optimize the following model prediction loss:
\begin{equation}
\small
    \mathcal{L}_{int} = \sum_{(\mathcal{G}^{1}, \mathcal{G}^{2}) \in \mathcal{D}}{\sum_{\Tilde{\mathcal{S}}^{1}}{\mathcal{L}(\mathbf{Y}, z_{\mathcal{C}^{1}}, z_{\mathcal{G}^{2}}, z_{\Tilde{\mathcal{S}}^{1}})}},
\end{equation}
where loss function $\mathcal{L}$ can be modeled as the cross entropy loss for classification and the root mean squared error loss for regression.
In practice, we make model predictions by concatenating $z_{\mathcal{C}^{1}}$ and $z_{\mathcal{G}^{2}}$ with randomly selected $z_{\tilde{\mathcal{S}}^{1}}$.

\subsubsection{Final Objectives.}
Finally, we train the model with the final objective function as follows:
\begin{equation}
\small
    \mathcal{L}_{final} = \mathcal{L}_{sup} + \mathcal{L}_{causal} + \lambda_{1} \cdot \mathcal{L}_{KL} + \lambda_{2} \cdot \mathcal{L}_{int},
\label{Eq: Final Objective}
\end{equation}
where $\lambda_{1}$ and $\lambda_{2}$ are weight hyperparameters for $\mathcal{L}_{KL}$ and $\mathcal{L}_{int}$, respectively.
Note that $\mathcal{L}_{sup}$ indicates the model loss given the pair of graph $(\mathcal{G}^{1}, \mathcal{G}^{2})$ and target $\mathbf{Y}$ without learning the causal substructure.



\section{Theoretical Analysis}

In this section, we provide theoretical analyses on the behavior of~\proposed~based on a concept in information theory, i.e., mutual information, which has made significant contributions in the various fields of machine learning \cite{alemi2016deep,hjelm2018learning,velickovic2019deep}.
Specifically, given a pair of molecules $(\mathcal{G}^{1}, \mathcal{G}^{2})$ in dataset $\mathcal{D}$, 
we assume that the molecule $\mathcal{G}^{1}$ contains an optimal causal substructure $\mathcal{C}^{1*}$ that is related to the causality of interaction behavior, i.e., $p(\mathbf{Y}|\mathcal{G}^{1}, \mathcal{G}^{2}) = p(\mathbf{Y}|\mathcal{C}^{1*}, \mathcal{G}^{2})$, where $p$ indicates the conditional distribution of label $\mathbf{Y}$.
We approximate $p$ with a hypothetical predictive model $q$, i.e., \proposed, by minimizing the negative conditional log-likelihood of label distribution as follows:
\begin{equation}
    - \ell = - \sum_{i = 1}^{n}{\log{q(\mathbf{Y}_{i}|\mathcal{C}^{1}_{i}, \mathcal{G}^{2}_{i})}}.
\end{equation}
This term can be expanded by multiplying and dividing $q$ with the term $p(\mathbf{Y}|\mathcal{C}^{1}, \mathcal{G}^{2})$ and $p(\mathbf{Y}|\mathcal{G}^{1}, \mathcal{G}^{2})$ as follows:
\begin{eqnarray}
\footnotesize
\begin{split}
    - \ell &= \sum_{i = 1}^{n}{\log {p(\mathbf{Y}_{i}|\mathcal{C}^{1}_{i}, \mathcal{G}^{2}_{i}) \over q(\mathbf{Y}_{i}|\mathcal{C}^{1}_{i}, \mathcal{G}^{2}_{i})}
    + \sum_{i = 1}^{n}{\log {p(\mathbf{Y}_{i}|\mathcal{G}^{1}_{i}, \mathcal{G}^{2}_{i}) \over p(\mathbf{Y}_{i}|\mathcal{C}^{1}_{i}, \mathcal{G}^{2}_{i})}}
    - \sum_{i = 1}^{n}{\log {p(\mathbf{Y}_{i}|\mathcal{G}^{1}_{i}, \mathcal{G}^{2}_{i})}}} \\
    &= \mathbb{E}\left[ \log{p(\mathbf{Y}|\mathcal{C}^{1}, \mathcal{G}^{2}) \over q(\mathbf{Y}|\mathcal{C}^{1}, \mathcal{G}^{2})}\right] 
    + \mathbb{E}\left[ \log {p(\mathbf{Y}|\mathcal{G}^{1}, \mathcal{G}^{2}) \over p(\mathbf{Y}|\mathcal{C}^{1}, \mathcal{G}^{2})} \right]
    - \mathbb{E} \left[ \log {p(\mathbf{Y}|\mathcal{G}^{1}, \mathcal{G}^{2})} \right],
\end{split}
\label{Eq: overall loss}
\end{eqnarray}
where $p(\mathbf{Y}|\mathcal{C}^{1}, \mathcal{G}^{2})$ and $p(\mathbf{Y}|\mathcal{G}^{1}, \mathcal{G}^{2})$ indicate the true label distributions given a pair of causal substructure $\mathcal{C}^{1}$ and molecule $\mathcal{G}^{2}$, and a pair of molecules $\mathcal{G}^{1}$ and $\mathcal{G}^{2}$, respectively.
Then, the second term can be solved as follows:
\begin{eqnarray}
\footnotesize
\begin{split}
    \mathbb{E}\left[ \log {p(\mathbf{Y}|\mathcal{G}^{1}_{i}, \mathcal{G}^{2}_{i}) \over p(\mathbf{Y}|\mathcal{C}^{1}_{i}, \mathcal{G}^{2}_{i})} \right] &= 
    \mathbb{E}\left[ \log {p(\mathbf{Y}|\mathcal{C}^{1}_{i}, \mathcal{S}^{1}_{i}, \mathcal{G}^{2}_{i}) \over p(\mathbf{Y}|\mathcal{C}^{1}_{i}, \mathcal{G}^{2}_{i})} \right] \\
    &= \sum_{i=1}^{n} {p(\mathcal{G}^{1}_{i}, \mathcal{G}^{2}_{i}, \mathbf{Y}_i) \log {p(\mathbf{Y}_{i}|\mathcal{C}^{1}_{i}, \mathcal{S}^{1}_{i}, \mathcal{G}^{2}_{i}) \over p(\mathbf{Y}_{i}|\mathcal{C}^{1}_{i}, \mathcal{G}^{2}_{i})}} \\
    &= \sum_{i=1}^{n} {p(\mathcal{G}^{1}_{i}, \mathcal{G}^{2}_{i}, \mathbf{Y}_i) \log {{p(\mathbf{Y}_{i}|\mathcal{C}^{1}_{i}, \mathcal{S}^{1}_{i}, \mathcal{G}^{2}_{i}) \over p(\mathbf{Y}_{i}|\mathcal{C}^{1}_{i}, \mathcal{G}^{2}_{i})}{p(\mathcal{S}^{1}_{i}|\mathcal{C}^{1}_{i}, \mathcal{G}^{2}_{i}) \over p(\mathcal{S}^{1}_{i}|\mathcal{C}^{1}_{i}, \mathcal{G}^{2}_{i})}}} \\
    &= \sum_{i=1}^{n} {p(\mathcal{G}^{1}_{i}, \mathcal{G}^{2}_{i}, \mathbf{Y}_i) \log {p(\mathcal{S}^{1}_{i}, \mathbf{Y}_{i}|\mathcal{C}^{1}_{i},  \mathcal{G}^{2}_{i}) \over p(\mathbf{Y}_{i}|\mathcal{C}^{1}_{i}, \mathcal{G}^{2}_{i})\cdot p(\mathcal{S}^{1}_{i}|\mathcal{C}^{1}_{i}, \mathcal{G}^{2}_{i})}} \\
    &= I(\mathcal{S}^{1}; \mathbf{Y}|\mathcal{C}^{1}, \mathcal{G}^{2})
\end{split}
\label{Eq: second term}
\end{eqnarray}
By plugging Equation \ref{Eq: second term} into Equation \ref{Eq: overall loss}, we have the following objective function for training \proposed:
\begin{equation}
    \min \mathbb{E}\left[ \log{p(\mathbf{Y}|\mathcal{C}^{1}, \mathcal{G}^{2}) \over q(\mathbf{Y}|\mathcal{C}^{1}, \mathcal{G}^{2})}\right] 
    + I(\mathcal{S}^{1}; \mathbf{Y}|\mathcal{C}^{1},  \mathcal{G}^{2})
    + H(\mathbf{Y}|\mathcal{G}^{1}, \mathcal{G}^{2}).
\end{equation}
The first term, which is the likelihood ratio between true distribution $p$ and predicted distribution $q$, will be minimized if the model can appropriately approximate the true distribution $p$ given causal substructure $\mathcal{C}^{1}$ and the paired molecule $\mathcal{G}^{2}$, and the third term is irreducible constant inherent in the dataset.
The second term, which is our main interest, indicates the conditional mutual information between the label information $\mathbf{Y}$ and shortcut substructure $\mathcal{S}^{1}$ given causal substructure $\mathcal{C}^{1}$ and the paired molecule $\mathcal{G}^{2}$.
Based on our derivation above, we explain the behavior of \proposed~in two different perspectives.

\begin{itemize}[leftmargin=2mm]
    \item \textbf{Perspective 1: \proposed~ learns informative causal substructure.}
The term $I(\mathcal{S}^{1}; \mathbf{Y}|\mathcal{C}^{1},  \mathcal{G}^{2})$ incentivizes the model to disentangle the shortcut substructure $\mathcal{S}^{1}$ that are no longer needed in predicting the label $\mathbf{Y}$ when the context $\mathcal{C}^{1}$ and $\mathcal{G}^{2}$ are given, which also aligns with the domain knowledge in molecular sciences, i.e., \textit{a certain functional group induces the same or similar chemical reactions regardless of other components that exist in the molecule}. 
Moreover, due to the chain rule of mutual information, i.e., $I(\mathcal{S}^{1}; \mathbf{Y}|\mathcal{C}^{1},  \mathcal{G}^{2}) = I(\mathcal{G}^{1}, \mathcal{G}^{2}; \mathbf{Y}) - I(\mathcal{C}^{1}, \mathcal{G}^{2}; \mathbf{Y})$, minimizing the second term encourages the causal substructure $\mathcal{C}^{1}$ and paired molecule $\mathcal{G}^{2}$ to contain enough information on target $\mathbf{Y}$.
Therefore, we argue that \proposed~learns to discover the causal substructure $\mathcal{C}^{1}$ that has enough information for predicting the target $\mathbf{Y}$ regarding the paired molecule $\mathcal{G}^{2}$, while ignoring the shortcut substructure $\mathcal{S}^{1}$ that will no longer provide useful information for the model prediction.

\item \textbf{Perspective 2: \proposed~reduces model bias with causal view.}
Besides, in the perspective of information leakage~\cite{dwork2012fairness,seo2022information}, it is possible to quantify the model bias based on mutual information. That is, the model bias is defined as the co-dependence between the shortcut substructure $\mathcal{S}^{1}$ and the target variable $\mathbf{Y}$, i.e., $I(\mathcal{S}^{1}; \mathbf{Y})$.
Therefore, to measure the model bias, we are solely interested in the direct path between $\mathcal{S}^{1}$ and $\mathbf{Y}$, i.e., $\mathcal{S}^{1} \rightarrow \mathcal{R}^{1} \rightarrow \mathbf{Y}$, in Figure \ref{fig2}.
However, there exist several backdoor paths induced by variables $\mathcal{C}^{1}$ and $\mathcal{G}^{2}$ which are inevitably correlated with $\mathcal{S}^{1}$.
Fortunately, such backdoor paths can be blocked by conditioning on confounding variables, i.e., $\mathcal{C}^{1}$ and $\mathcal{G}^{2}$, enabling the direct measure of the model bias via conditional mutual information $I(\mathcal{S}^{1}; \mathbf{Y}|\mathcal{C}^{1}, \mathcal{G}^{2})$.
Therefore, we argue that \proposed~learns to minimize the model bias with conditional mutual information term.
\end{itemize}

\section{Experiments}
\subsection{Experimental Setup}
\subsubsection{Datasets.}

We use \textbf{fourteen} real-world datasets and a synthetic dataset to comprehensively evaluate the performance of \proposed~on three tasks, i.e., 1) molecular interaction prediction, 2) drug-drug interaction (DDI) prediction, and 3) graph similarity learning.
Specifically, for molecular interaction prediction, we use a dataset related to optical and photophysical properties of chromophore with various solvents, i.e., \textbf{Chromophore} dataset \cite{Chromophore}, and five datasets related to solvation free energy of the solute with various solvents, i.e., \textbf{MNSol} \cite{MNSol}, \textbf{FreeSolv} \cite{FreeSolv}, \textbf{CompSol} \cite{CompSol}, \textbf{Abraham} \cite{Abraham}, and \textbf{Combisolv} \cite{CombiSolv}.
In Chromophore dataset, we use maximum absorption wavelength (\textbf{Absorption}), maximum emission wavelength (\textbf{Emission}), and excited state lifetime (\textbf{Lifetime}) properties.
For DDI prediction task, we use three datasets, i.e., \textbf{ZhangDDI} \cite{ZhangDDI}, \textbf{ChChMiner} \cite{ChChMiner}, and \textbf{DeepDDI} \cite{DeepDDI}, all of which contain side-effect information on taking two drugs simultaneously.
Moreover, we use five datasets for graph similarity learning, i.e., \textbf{AIDS}, \textbf{LINUX}, \textbf{IMDB} \cite{bai2019simgnn}, \textbf{FFmpeg}, and \textbf{OpenSSL} \cite{xu2017neural,ling2019hierarchical}, containing the similarity information between two graph structures.
The detailed statistics and descriptions are given in Appendix \ref{app: Datasets}.

\begin{table*}[t]
\begin{minipage}{0.75\linewidth}{
    \centering
    \small
    \caption{Performance on molecular interaction prediction task (regression).}
    \vspace{-2ex}
    \resizebox{0.99\linewidth}{!}{
    \begin{tabular}{c|ccc|ccccc}
               & \multicolumn{3}{c|}{Chromophore}            & \multirow{2}{*}{MNSol} & \multirow{2}{*}{FreeSolv} & \multirow{2}{*}{CompSol} & \multirow{2}{*}{Abraham} & \multirow{2}{*}{CombiSolv} \\ \cline{2-4} 
               & Absorption   & Emission     & Lifetime      & & & & & \\ \hline \hline
    GCN        & $25.75 $ \scriptsize{(1.48)} & $31.87 $ \scriptsize{(1.70)} & $0.866 $ \scriptsize{(0.015)} & $0.675 $ \scriptsize{(0.021)} & $1.192 $ \scriptsize{(0.042)} & $0.389 $ \scriptsize{(0.009)} & $0.738 $ \scriptsize{(0.041)} & $0.672 $ \scriptsize{(0.022)} \\
    GAT        & $26.19 $ \scriptsize{(1.44)} & $30.90 $ \scriptsize{(1.01)} & $0.859 $ \scriptsize{(0.016)} & $0.731 $ \scriptsize{(0.007)} & $1.280 $ \scriptsize{(0.049)} & $0.387 $ \scriptsize{(0.010)} & $0.798 $ \scriptsize{(0.038)} & $0.662 $ \scriptsize{(0.021)} \\
    MPNN       & $24.43 $ \scriptsize{(1.55)} & $30.17 $ \scriptsize{(0.99)} & $0.802 $ \scriptsize{(0.024)} & $0.682 $ \scriptsize{(0.017)} & $1.159 $ \scriptsize{(0.032)} & $0.359 $ \scriptsize{(0.011)} & $0.601 $ \scriptsize{(0.035)} & $0.568 $ \scriptsize{(0.005)} \\
    GIN        & $24.92 $ \scriptsize{(1.67)} & $32.31 $ \scriptsize{(0.26)} & $0.829 $ \scriptsize{(0.027)} & $0.669 $ \scriptsize{(0.017)} & $1.015 $ \scriptsize{(0.041)} & $0.331 $ \scriptsize{(0.016)} & $0.648 $ \scriptsize{(0.024)} & $0.595 $ \scriptsize{(0.014)} \\ \hline
    CIGIN      & $19.32 $ \scriptsize{(0.35)} & $25.09 $ \scriptsize{(0.32)} & $0.804 $ \scriptsize{(0.010)} & $0.607 $ \scriptsize{(0.024)} & $0.905 $ \scriptsize{(0.014)} & $0.308 $ \scriptsize{(0.018)} & $0.411 $ \scriptsize{(0.008)} & $0.451 $ \scriptsize{(0.009)} \\ \hline
    \proposed       & $\mathbf{17.93}$ \scriptsize{(0.31)} & $\mathbf{24.30}$ \scriptsize{(0.22)} & $\mathbf{0.776}$ \scriptsize{(0.007)} & $\mathbf{0.551}$ \scriptsize{(0.017)}  & $\mathbf{0.815}$ \scriptsize{(0.046)} & $\mathbf{0.255}$ \scriptsize{(0.011)} & $\mathbf{0.374}$ \scriptsize{(0.011)} & $\mathbf{0.421}$ \scriptsize{(0.008)} \\ \hline
    \end{tabular}}
    \label{tab: regression}
}\end{minipage}
\begin{minipage}{0.23\linewidth}{
    \centering
    \includegraphics[width=0.85\linewidth]{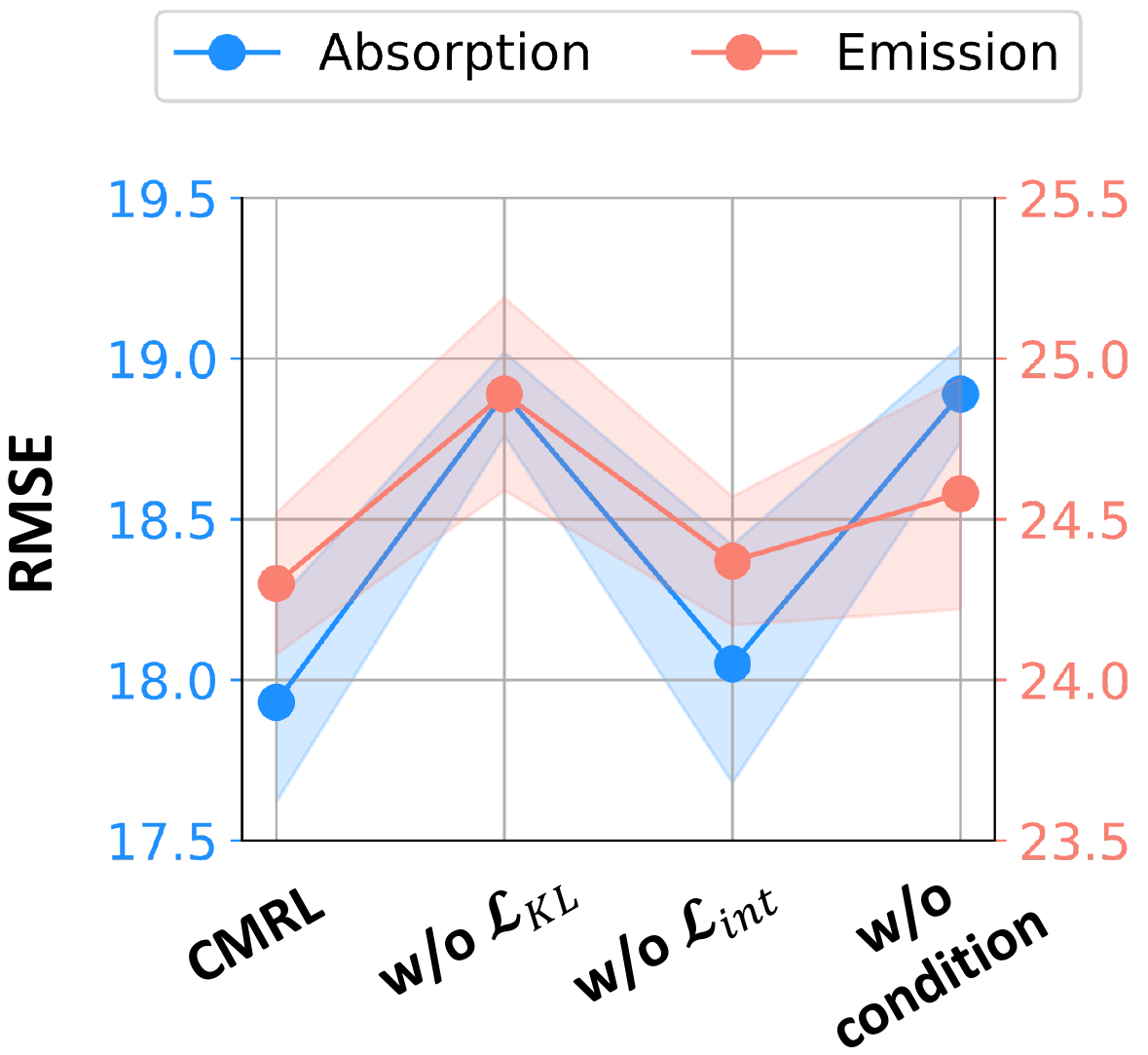}
    \vspace{-2ex}
    \captionof{figure}{Model ablations.}
    \label{fig: Ablations}
}\end{minipage}
\vspace{-3ex}
\end{table*}

\subsubsection{Methods Compared.}
We compare \proposed~with the state-of-the-art methods in each task.
Specifically, we mainly compare \proposed~with CIGIN \cite{pathak2020chemically} for molecular interaction task.
For DDI prediction task, we mainly compare with SSI-DDI \cite{nyamabo2021ssi} and MIRACLE \cite{wang2021multi}, and additionally compare with CIGIN \cite{pathak2020chemically} by changing the prediction head that was originally designed for regression to classification.
Moreover, for both tasks, we include simple baseline methods, i.e., GCN \cite{kipf2016semi}, GAT \cite{velivckovic2017graph}, MPNN \cite{gilmer2017neural}, and GIN \cite{xu2018powerful}. 
Specifically, we independently encode a pair of graphs and concatenate two embeddings to predict the target value with an MLP predictor.
For the similarity learning task, we mainly compare \proposed~with $\text{H}^{2}\text{MN}$ \cite{zhang2021h2mn}, and also compare with the performance of SimGNN \cite{bai2019simgnn}, GMN \cite{li2019graph}, GraphSim \cite{bai2020learning}, and HGMN \cite{ling2019hierarchical} reported in $\text{H}^{2}\text{MN}$.

\subsubsection{Evaluation Protocol.}
For the molecular interaction prediction task, we evaluate the models under 5-fold cross validation scheme following previous work \cite{pathak2020chemically}.
Specifically, the dataset is randomly split into 5 subsets and one of the subsets is used as the test set while the remaining subsets are used to train the model.
Moreover, a subset of the test set is used as a validation set for hyperparameter selection and early stopping.
Finally, we report the mean and standard deviation of 5 repeats of 5-fold cross validation (i.e., 25 runs in total).
For DDI prediction task, we conduct experiments on two data split settings, i.e., in-distribution and out-of-distribution.
Specifically, for in-distribution setting, we split the data into train/valid/test data of 60/20/20$\%$ following previous work \cite{nyamabo2021ssi}, whose drugs in the test set are also included in the training set.
On the other hand, in the out-of-distribution setting, we assess the model's performance on molecules belonging to new scaffold classes that were not part of the training dataset. 
As shown in Appendix \ref{app: scaffold split for ood}, this introduces a distribution shift, as the model encounters novel molecular structures during evaluation.
Specifically, let $\mathbb{G}$ denote the total set of molecules in the dataset and assume that there exist a total number of $s$ distinct scaffolds in the dataset.
Then, the scaffold classes consist of drug sets $\mathbb{G}_{1},\ldots, \mathbb{G}_{s}$, where $\mathbb{G}_{k}$ contains all drugs in $\mathbb{G}$ that belong to the $k$-th scaffold.
We construct a set of training drugs $\mathbb{G}_{\mathrm{ID}}$ with $c$ classes of scaffold, i.e., $\mathbb{G}_{1},\ldots, \mathbb{G}_{c}$, and a set of test drugs $\mathbb{G}_{\mathrm{OOD}}$ with the remain classes of scaffold, i.e., $\mathbb{G}_{c+1},\ldots, \mathbb{G}_{s}$.
Then, the new split of the drug pair dataset consists of $\mathcal{D}_{\mathrm{ID}} = \{ (\mathcal{G}^{1}, \mathcal{G}^{2}) \in \mathcal{D} | \mathcal{G}^{1} \in \mathbb{G}_{\mathrm{ID}} \wedge \mathcal{G}^{2} \in \mathbb{G}_{\mathrm{ID}} \}$ and $\mathcal{D}_{\mathrm{OOD}} = \{ (\mathcal{G}^{1}, \mathcal{G}^{2}) \in \mathcal{D} | (\mathcal{G}^{1} \in \mathbb{G}_{\mathrm{OOD}} \wedge \mathcal{G}^{2} \in \mathbb{G}_{\mathrm{OOD}}) \vee (\mathcal{G}^{1} \in \mathbb{G}_{\mathrm{OOD}} \wedge \mathcal{G}^{2} \in \mathbb{G}_{\mathrm{ID}}) \vee (\mathcal{G}^{1} \in \mathbb{G}_{\mathrm{ID}} \wedge \mathcal{G}^{2} \in \mathbb{G}_{\mathrm{OOD}}) \}$, whose subset was used as validation set.
For both the in-distribution and out-of-distribution settings, we repeat 5 experiments with different random seeds on a split, and report the mean and standard deviation of the repeats.
For the similarity learning task, we repeat 5 independent experiments with different random seeds on the already-split data given by \cite{zhang2021h2mn}.
For all three tasks, we report the test performance when the performance on the validation set gives the best result.

\subsubsection{Evaluation Metrics.}
We evaluate the performance of \proposed~in terms of RMSE for molecular interaction prediction, AUROC, and Accuracy for DDI prediction, and MSE, Spearman's Rank Correlation Coefficient ($\rho$), precision@10 (p@10), and AUROC for graph similarity learning.

\subsubsection{Implementation Details.}
For the molecular interaction prediction task, we use a 3-layer MPNN \cite{gilmer2017neural} molecular encoder and a 3-layer MLP predictor with ReLU activation following previous work \cite{pathak2020chemically}.
For DDI prediction task, we use a 3-layer GIN \cite{xu2018powerful} drug encoder and a single layer MLP predictor without activation.
For the graph similarity learning task, we use a 3-layer GCN \cite{kipf2016semi} graph encoder and a 3-layer MLP predictor with ReLU activation following previous work \cite{zhang2021h2mn}.
More details on the model implementation and hyperparameter specifications are given in Appendix \ref{App: Implementation Details}.

\vspace{-1ex}
\subsection{Overall Performance}

\begin{table*}[t]
    \centering
    \small
    \caption{Performance on drug-drug interaction prediction task (classification).}
    \vspace{-2ex}
    \resizebox{0.99\linewidth}{!}{
    \begin{tabular}{c|cccccc|cccccc}
        & \multicolumn{6}{c}{(a) In-Distribution} & \multicolumn{6}{|c}{(b) Out-of-Distribution} \\ \cline{2-13}
        & \multicolumn{2}{c}{ZhangDDI} & \multicolumn{2}{c}{ChChMiner} & \multicolumn{2}{c}{DeepDDI} & \multicolumn{2}{|c}{ZhangDDI} & \multicolumn{2}{c}{ChChMiner} & \multicolumn{2}{c}{DeepDDI} \\ \cline{2-13} 
        & AUROC & Accuracy & AUROC & Accuracy & AUROC & Accuracy & AUROC & Accuracy & AUROC & Accuracy & AUROC & Accuracy \\ \hline \hline
        GCN        & $91.64 $ \scriptsize{(0.31)} & $83.31 $ \scriptsize{(0.61)} & $94.71 $ \scriptsize{(0.33)} & $87.36 $ \scriptsize{(0.24)} & $92.02 $ \scriptsize{(0.01)} & $86.96 $ \scriptsize{(0.02)} & $70.61$ \scriptsize{(2.32)} & $64.22$ \scriptsize{(1.64)} & $74.17 $ \scriptsize{(0.89)} & $67.56 $ \scriptsize{(1.29)} & $76.38 $ \scriptsize{(0.43)} & $67.92 $ \scriptsize{(0.81)}\\
        GAT        & $92.10 $ \scriptsize{(0.28)} & $84.14 $ \scriptsize{(0.38)} & $96.15 $ \scriptsize{(0.53)} & $89.49 $ \scriptsize{(0.88)} & $92.01 $ \scriptsize{(0.02)} & $86.99 $ \scriptsize{(0.05)} & $73.15 $ \scriptsize{(2.50)} & $65.14 $ \scriptsize{(2.47)} & $75.64 $ \scriptsize{(0.99)} & $68.61 $ \scriptsize{(0.72)} & $76.44 $ \scriptsize{(1.27)} & $67.94 $ \scriptsize{(1.38)} \\
        MPNN       & $92.34 $ \scriptsize{(0.35)} & $84.56 $ \scriptsize{(0.31)} & $96.25 $ \scriptsize{(0.53)} & $90.02 $ \scriptsize{(0.42)} & $92.02 $ \scriptsize{(0.02)} & $86.97 $ \scriptsize{(0.01)} & $72.39 $ \scriptsize{(1.70)} & $64.55 $ \scriptsize{(1.75)} & $76.40 $ \scriptsize{(0.91)} & $68.51 $ \scriptsize{(0.71)} & $79.03 $ \scriptsize{(0.81)} & $71.23 $ \scriptsize{(0.90)}  \\
        GIN        & $93.16 $ \scriptsize{(0.04)} & $85.59 $ \scriptsize{(0.05)} & $97.52 $ \scriptsize{(0.05)} & $91.89 $ \scriptsize{(0.66)} & $92.03 $ \scriptsize{(0.00)} & $87.02 $ \scriptsize{(0.03)} & $75.04 $ \scriptsize{(0.63)} & $67.14 $ \scriptsize{(1.03)} & $74.32 $ \scriptsize{(2.93)} & $67.49 $ \scriptsize{(2.44)} & $78.61 $ \scriptsize{(0.58)} & $70.33 $ \scriptsize{(1.11)}  \\ \hline
        MIRACLE    & $93.05 $ \scriptsize{(0.07)} & $84.90 $ \scriptsize{(0.36)} & $88.66 $ \scriptsize{(0.37)} & $84.29 $ \scriptsize{(0.14)} & 62.23 \scriptsize{(0.75)} & 62.35 \scriptsize{(0.30)} & 59.57 \scriptsize{(0.90)} & 52.31 \scriptsize{(2.24)} & $73.28$ \scriptsize{(0.71)} & $50.49$ \scriptsize{(0.59)} & 62.32 \scriptsize{(1.63)} & 51.30 \scriptsize{(0.29)} \\ 
        SSI-DDI    & $92.74 $ \scriptsize{(0.12)} & $84.61 $ \scriptsize{(0.18)} & $98.44 $ \scriptsize{(0.08)} & $93.50 $ \scriptsize{(0.16)} & $93.97 $ \scriptsize{(0.38)} & $88.44 $ \scriptsize{(0.39)} & $71.67 $ \scriptsize{(4.71)} & $65.78 $ \scriptsize{(3.02)} & $75.59 $ \scriptsize{(1.93)} & $68.75 $ \scriptsize{(1.41)} & $80.41 $ \scriptsize{(1.74)} & $72.05 $ \scriptsize{(1.47)}  \\
        CIGIN      & $93.28 $ \scriptsize{(0.13)} & $85.54 $ \scriptsize{(0.30)} & $98.51 $ \scriptsize{(0.10)} & $93.77 $ \scriptsize{(0.25)} & $99.12 $ \scriptsize{(0.03)} & $96.55 $ \scriptsize{(0.11)} & $73.99 $ \scriptsize{(1.74)} & $66.44 $ \scriptsize{(1.07)} & $80.24 $ \scriptsize{(2.00)} & $73.28 $ \scriptsize{(1.08)} & $83.78 $ \scriptsize{(0.87)} & $74.07 $ \scriptsize{(1.19)} \\ \hline
        \proposed       & $\mathbf{93.73}$ \scriptsize{(0.15)} & $\mathbf{86.32 }$ \scriptsize{(0.23)} & $\mathbf{98.70}$ \scriptsize{(0.05)} & $\mathbf{94.26}$ \scriptsize{(0.28)} & $\mathbf{99.13} $ \scriptsize{(0.02)} & $\mathbf{96.70} $ \scriptsize{(0.12)} & $\mathbf{75.30} $ \scriptsize{(1.39)} & $\mathbf{67.76} $ \scriptsize{(1.41)} & $\mathbf{82.05} $ \scriptsize{(0.67)} & $\mathbf{74.21} $ \scriptsize{(0.78)} & $\mathbf{83.83} $ \scriptsize{(0.97)} & $\mathbf{75.20} $ \scriptsize{(0.66)} \\ \hline
    \end{tabular}}
    \label{tab: classification}
    \vspace{-2ex}
\end{table*}

\begin{table*}[t]
\begin{minipage}{0.7\linewidth}{
    \centering
    \small
    \caption{Performance on similarity learning task (regression/classification).}
    \vspace{-2ex}
    \resizebox{0.99\linewidth}{!}{
    \begin{tabular}{c|ccc|ccc|ccc|c|c}
               & \multicolumn{3}{c|}{AIDS} & \multicolumn{3}{c|}{LINUX} & \multicolumn{3}{c|}{IMDB} & FFmpeg & OpenSSL \\ \cline{2-12}
               & MSE & $\rho$ & p@10 & MSE & $\rho$ & p@10 & MSE & $\rho$ & p@10 & AUROC & AUROC \\ \hline \hline
    SimGNN     & $1.376$ & $0.824$ & $0.400$ & $2.479$ & $0.912$ & $0.635$ & $1.264$ & $0.878$ & $0.759$ & $93.45$ & $94.25$\\
    GMN        & $4.610$ & $0.672$ & $0.200$ & $2.571$ & $0.906$ & $0.888$ & $4.422$ & $0.725$ & $0.604$ & $94.76$ & $93.91$\\
    GraphSim   & $1.919$ & $0.849$ & $0.446$ & $0.471$ & $0.976$ & $0.956$ & $0.743$ & $0.926$ & $0.828$ & $94.48$ & $93.66$ \\
    HGMN       & $1.169$ & $\mathbf{0.905}$ & $0.456$ & $0.439$ & $0.985$ & $0.955$ & $0.335$ & $0.919$ & $0.837$& $97.83$ & $95.87$ \\ \hline
    $\text{H}^2\text{MN}_{\text{RW}}$ & $0.936$ & $0.878$ & $0.496$ & $0.136$ & $0.988$ & $0.970$& $0.296$ & $0.918$ & $0.872$ & $\mathbf{99.05}$ & $92.21$ \\
    $\text{H}^2\text{MN}_{\text{NE}}$ & $0.924$ & $0.883$ & $0.511$ & $0.130$ & $0.990$ & $0.978$& $0.297$ & $0.889$ & $0.875$ & $98.16$ & $\mathbf{98.25}$ \\ \hline
    \proposed & $\mathbf{0.770}$ & $0.899$ & $\mathbf{0.574}$ & $\mathbf{0.094}$ & $\mathbf{0.992}$ & $\mathbf{0.989}$ & $\mathbf{0.263}$ & $\mathbf{0.944}$ & $\mathbf{0.879}$ & $98.69$ & $96.57$ \\ \hline
    \end{tabular}}
    \label{tab:similarity learning}
} \end{minipage}
\begin{minipage}{0.29\linewidth}{
\centering
    \includegraphics[width=0.75\linewidth]{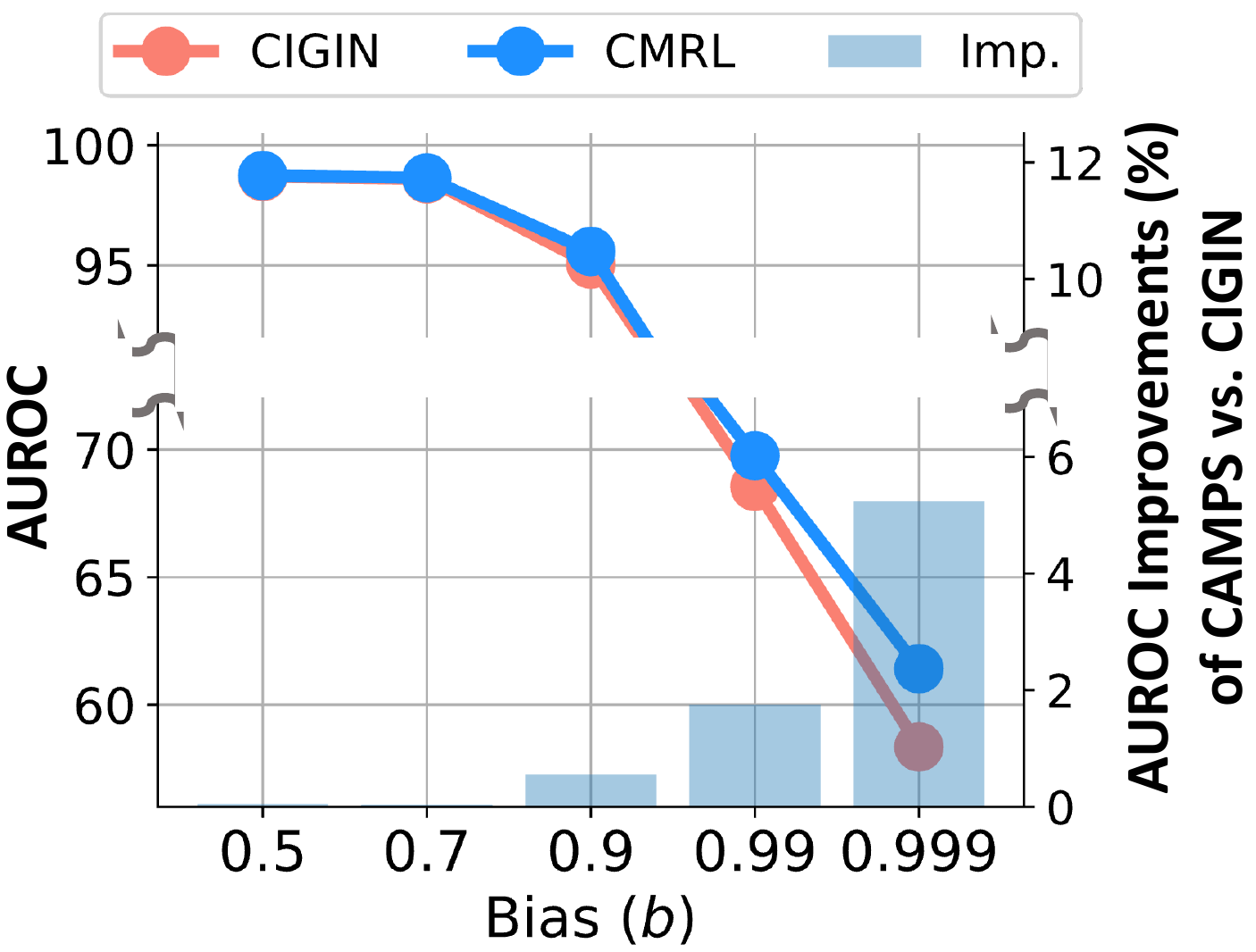} 
    \vspace{-2ex}
    \captionof{figure}{Performance on a synthetic dataset with various bias levels.}
    \label{fig:biased dataset}   
}\end{minipage}
\vspace{-2ex}
\end{table*}

The experimental results on fourteen datasets on three distinctive tasks, i.e., molecular interaction prediction, DDI prediction, and graph similarity learning, are presented in Table \ref{tab: regression}, Table \ref{tab: classification}, and Table \ref{tab:similarity learning}, respectively. We have the following observations:
\textbf{1)} \proposed~outperforms all other baseline methods that overlook the significance of the causal substructure during training, i.e., CIGIN, MIRACLE, and SSI-DDI, in both molecular interaction prediction and DDI prediction tasks.
We argue that learning the causal substructure, which is causally related to the target prediction during the decision process, incentivizes the model to learn from the minimal but informative component of molecules, e.g., the functional group, thereby improving the generalization capability of the model.
Considering the domain knowledge in molecular sciences, i.e., a functional group undergoes the same or comparable chemical reactions regardless of the remaining part of a molecule~\cite{smith2020march,mcnaught1997compendium}, it is natural to consider the causal substructure in molecular relational learning tasks.
\textbf{2)} A further benefit of learning the causal substructure is that \proposed~outperforms previous works on out-of-distribution scenarios for DDI prediction (see Table \ref{tab: classification} (b)), which is more challenging but more practical than in-distribution scenarios (see Table \ref{tab: classification} (a)).
This can also be explained by the generalization ability of \proposed, whose knowledge learned from the causal substructure (i.e., functional group) can be transferred to the chemicals consisting of the same causal substructure and the new classes of scaffolds.
This transferability of knowledge enables the model to remain robust in the face of distribution shifts within the chemical domain.
\textbf{3)} On the other hand, it is worth noting that simple baseline methods whose predictions are made with simple concatenated representations of molecules, i.e., GCN, GAT, MPNN, and GIN, perform worse than the methods that explicitly model the interaction between the atoms or substructures, i.e., SSI-DDI and CIGIN.
This indicates that modeling the interaction behavior between the small components of molecules is crucial for molecular relational learning tasks.
\textbf{4)} To demonstrate the wide applicability of \proposed~on various tasks, we conduct experiments on the graph similarity learning task, which aims to learn the similarity between a given pair of graphs (Table~\ref{tab:similarity learning}).
We observe that \proposed~outperforms all the baseline methods that are specifically designed for the similarity learning task in AIDS, LINUX, and IMDB datasets, demonstrating the applicability of \proposed~not only for the molecular relational learning but also for general graph relational learning tasks.
By comparing \proposed~to $\text{H}^{2}\text{MN}$, which also discovers multiple significant substructures based on Personalize PageRank (PPR) score \cite{page1999pagerank,haveliwala2002topic} but does not consider the causal substructure regarding paired graph,
we argue that discovering causal substructure regarding the paired graph is also crucial for general graph relational learning tasks.
On the other hand, we observe that \proposed~performs competitively on FFmpeg and OpenSSL datasets.
We attribute this to the inherent characteristics of datasets, whose graphs are control flow graphs of binary function, making it hard for the model to determine the causal substructure.
However, considering that most graph structures in the real world contain causal substructure, \proposed~can be widely adopted to general graph structures.

\vspace{-1ex}
\subsection{Experiments on Synthetic Dataset}
\label{sec: Performance on Synthetic Dataset}
In this section, we conduct experiments on a synthetic dataset with various bias levels to verify the robustness of \proposed~on the biased dataset compared with CIGIN \cite{pathak2020chemically}, which can be recognized as \proposed~without the causal framework.
To do so, we first construct synthetic graphs following \cite{sui2022causal,ying2019gnnexplainer}, each of which contains one of four causal substructures, i.e., ``House'', ``Cycle'', ``Grid'', and ``Diamond''.
Moreover, each graph also contains one of the shortcut substructures, i.e., ``Tree'' and ``BA.''
Then, our model learns to classify whether a pair of synthetic graphs contains the same causal substructure or not, i.e., binary classification.
We define the bias level $b$ of the dataset as the ratio of the positive pairs containing ``BA.'' shortcut substructures, while defining the portion of the negative pairs containing ``BA.'' shortcut substructures as $1-b$.
By doing so, as the bias level $b$ gets larger, ``BA.'' substructures will be the dominating substructure in model prediction.
More detail on the synthetic data generation process is provided in Appendix \ref{App: Synthetic Dataset Generation}.
In Figure \ref{fig:biased dataset}, we observe that the models' performance degrades as the bias of the dataset gets severe.
This indicates that as the bias gets severe, ``BA.'' shortcut substructure is more likely to confound the models to make predictions based on the shortcut substructure.
On the other hand, the performance gap between \proposed~and CIGIN gets larger as the bias gets severe.
This is because \proposed~learns the causality between the casual substructure and the target variable, which enables the model to alleviate the effect of the biased dataset on the model performance.
Since real-world datasets inevitably contain bias due to their data collection process \cite{sui2022causal,fan2022debiasing,fan2021generalizing}, we argue that \proposed~can be successfully adopted in real-world applications.

\begin{table*}[t]
\begin{minipage}{0.35\linewidth}{
    \centering
    \includegraphics[width=0.99\linewidth]{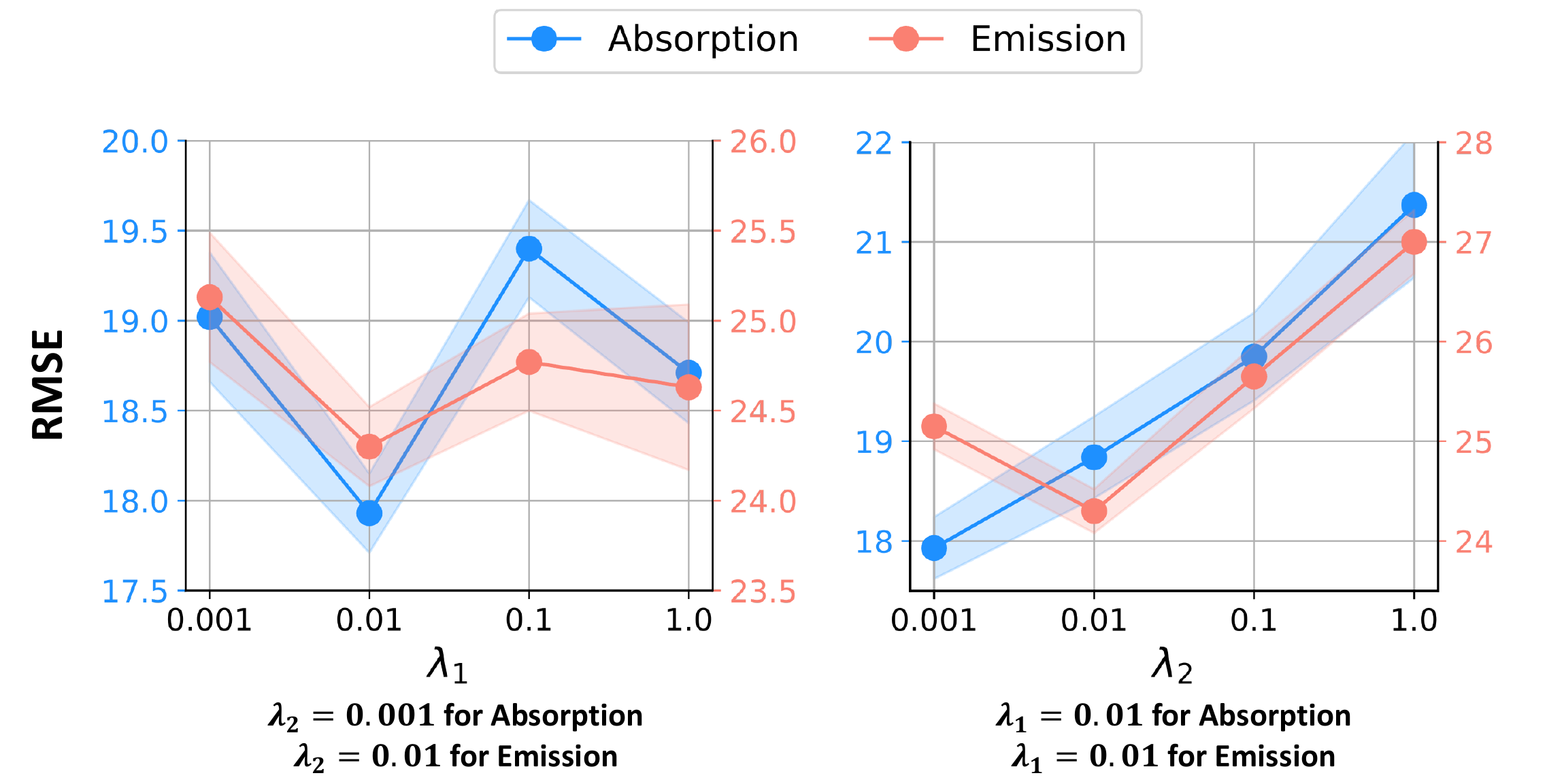}
    \vspace{-4ex}
    \captionof{figure}{Sensitivity analysis.}
    \label{fig: Sensitivity} 
} \end{minipage}
\begin{minipage}{0.64\linewidth}{
    \centering
    \includegraphics[width=0.85\linewidth]{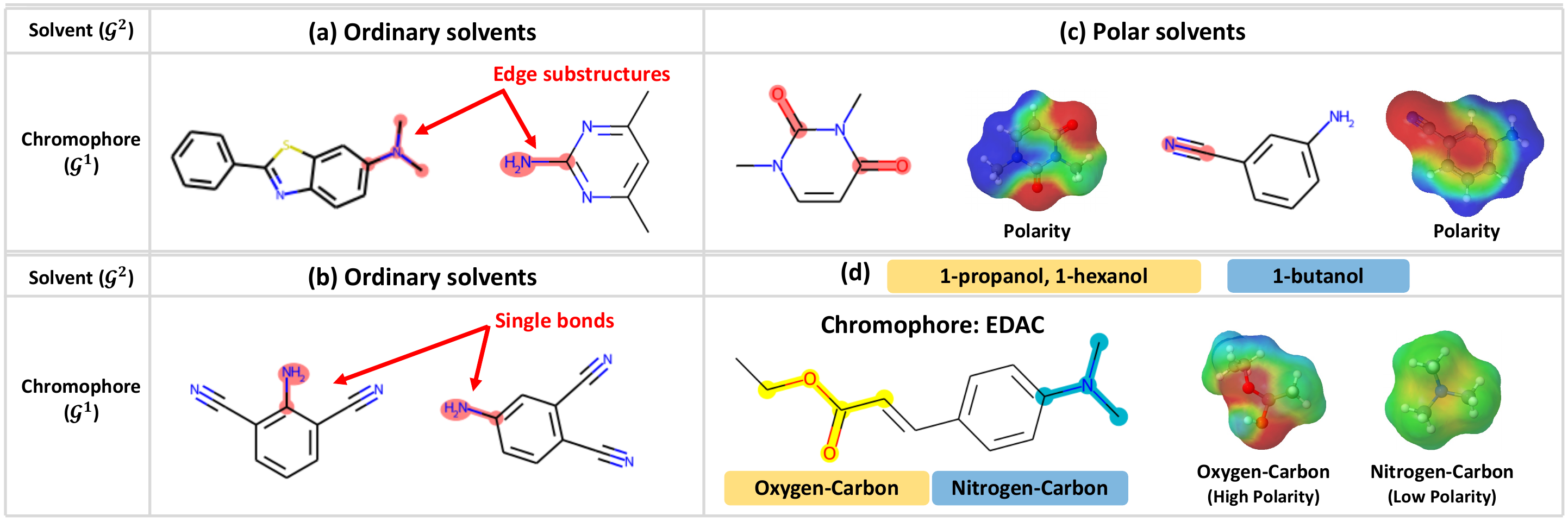} 
    \vspace{-2ex}
    \captionof{figure}{Qualitative analysis result. (Red: Predicted causal substructure)}
    \label{fig: qualitative analysis}  
}\end{minipage}
\vspace{-2ex}
\end{table*}

\vspace{-1ex}
\subsection{Model Analysis}
\subsubsection{Ablation Studies.}
\label{sec: Ablation Studies}
To demonstrate the importance of each component in \proposed, we conduct ablation studies as shown in Figure \ref{fig: Ablations}.
We have the following observations:
\textbf{1)} Disentangling the causal substructure $\mathcal{C}^{1}$ and the shortcut substructure $\mathcal{S}^{1}$ is crucial for causal intervention, as it enables the model to not only learn the causal substructure but also provides various confounders.
Therefore, the model without the disentanglement (i.e., w/o $\mathcal{L}_{KL}$) cannot properly learn causality.
\textbf{2)} On the other hand, we observe that training the model with disentangled causal substructures without causal intervention (i.e., w/o $\mathcal{L}_{int}$) performs competitively to \proposed.
This implies that learning with the causal substructure is crucial in molecular relational learning tasks, which enables the model to learn from the basic component in molecules that determines various chemical reactions.
\textbf{3)} What's interesting is that the model that naively models the intervention (i.e., w/o condition), whose confounders are not conditioned on the paired molecule $\mathcal{G}^{2}$, performs much worse than the model without intervention (i.e., w/o $\mathcal{L}_{int}$).
We attribute this to the wider variety of confounders than is required by the model, i.e., the wideness of the intervention space, which might give noisy signals to the model during the training.
\textbf{4)} On the other hand, \proposed, whose intervention is done by conditioning the confounders on the paired graphs, outperforms the model without intervention (i.e., w/o $\mathcal{L}_{int}$), indicating that \proposed~successfully adopts causal inference in molecular relational learning tasks.
We argue that a key to the success of \proposed~is conditional intervention, which generates the virtual confounders conditioned on the paired molecule $\mathcal{G}^{2}$, thereby preventing the model from learning from the unnecessary interfering confounders.

\subsubsection{Sensitivity Analysis.}
\label{sec: Sensitivity Analysis}
In Figure \ref{fig: Sensitivity}, we analyze the impact of hyperparameters for model training, i.e., $\lambda_{1}$ and $\lambda_{2}$, which control the strength of the disentangling loss $\mathcal{L}_{KL}$ and the effect of the intervention loss $\mathcal{L}_{int}$ in our final objectives in Equation \ref{Eq: Final Objective}, respectively.
We have the following observations:
\textbf{1)} The model consistently performs the worst when $\lambda_{2} = 1.0$, i.e., when the predictions made with confounders are treated equally to that made without confounders. 
This is because the model is seriously exposed to the noise induced by the confounders.
\textbf{2)} On the other hand, decreasing $\lambda_{2}$ does not always lead to good performance (also see Figure \ref{fig: Ablations}).
This is because the intervention loss $\mathcal{L}_{int}$ encourages the model to make a robust prediction with a causal substructure with various confounders.
Putting two observations together, it is crucial to find the optimal values of $\lambda_{2}$ for building a robust model.
\textbf{3)} However, $\lambda_{1}$ does not show a certain relationship with the model performance, indicating that the optimal value of the coefficient $\lambda_{1}$ should be tuned.


\subsection{Qualitative Analysis}
\label{sec: Qualitative Analysis}
In this section, we qualitatively analyze the causal substructure selected by \proposed~based on the chemical domain knowledge regarding the strength of the bonds and polarity of the molecules.
We have the following observations:
\textbf{1)} As shown in Figure \ref{fig: qualitative analysis} (a), our model selects the edge substructures in chromophores that are important for predicting the chromophore-solvent reaction.
This aligns with the chemical domain knowledge that chemical reactions usually happen around ionized atoms \cite{hynes1985chemical}.
\textbf{2)} On the other hand, among the various edge substructures, we find out that the model concentrates more on single-bonded substructures than triple-bonded substructures in Figure \ref{fig: qualitative analysis} (b).
This also aligns with the chemical domain knowledge that single-bonded substructures are more likely to undergo chemical reactions than triple-bonded substructures due to their weakness.
\textbf{3)} Moreover, when the chromophores react with polar solvents in Figure \ref{fig: qualitative analysis} (c), e.g., acetonitrile and ethanol, we observe that the model focuses on the edge substructures of high polarity\footnote{\label{footnote: polarity} The polarity of the chromophores and substructures is denoted in the right side of each chromophore in Figure \ref{fig: qualitative analysis}(c) and (d), respectively. As colors differ in a molecule, its polarity gets higher.}.
Considering that the polar solvents usually interact with polar molecules \cite{reichardt1965empirical}, edge substructures of high polarity are more likely to undergo chemical reactions when interacting with polar solvents.
\textbf{4)} One interesting observation here is that even with the same chromophore, the selected important substructure varies as the solvent varies.
Specifically, Figure \ref{fig: qualitative analysis} (d) shows the important substructure in the chromophore named trans-ethyl p-(dimethylamino) cinamate (EDAC) \cite{singh2009effect} detected by \proposed~when EDAC interacts with three different solvents: 1-propanol, 1-butanol, and 1-hexanol.
We observe that the model predicts oxygen-carbon substructure (marked in yellow) to be important when EDAC interacts with the 1-propanol solvent, while nitrogen-carbon substructure  (marked in blue) is predicted to be important when EDAC interacts with the 1-butanol solvent.
This can also be explained in terms of the polarity, where the high polarity solvent, i.e., 1-propanol, reacts with high-polarity substructure, i.e., oxygen-carbon substructure, while the low polarity solvent, i.e., 1-butanol, reacts with low-polarity substructure, i.e., nitrogen-carbon substructure\textsuperscript{\ref{footnote: polarity}}.
On the other hand, in the case of 1-hexanol, which is widely known as a non-polar solvent, \proposed~predicts oxygen-carbon substructure, which has a high polarity, to be important due to the local polarity of OH substructures in 1-hexanol.
The qualitative analysis demonstrates the explainability of \proposed, which can further accelerate the discovery process of molecules with similar properties.


\section{Conclusion}
In this paper, we propose a novel framework \proposed~for molecular relational learning tasks inspired by a well-known domain knowledge in molecular sciences, i.e., substructures in molecules such as a functional group determines the chemical reaction of a molecule regardless of other remaining atoms in the molecule.
The main idea is to discover the causal substructure in a molecule that has a true causality to the model prediction regarding a paired molecule with a novel conditional intervention framework.
By doing so, \proposed~learns from the minimal but informative component of molecules, improving the generalization capability of the model in various real-world scenarios.
Extensive experiments demonstrate the superiority of \proposed~in not only various molecular relational learning tasks but also general graph relational learning tasks. \\[3pt]

\noindent \textbf{Acknowledgement.}
This work was supported by the National Research Foundation of Korea (NRF) grant funded by the Korea government (MSIT) (No.2021R1C1C1009081), and Institute of Information \& communications Technology Planning \& Evaluation (IITP) grant funded by the Korea government (MSIT) (No.2022-0-00077).

\clearpage
\bibliographystyle{ACM-Reference-Format}
\balance
\bibliography{CMRL}

\clearpage

\appendix
\section{Appendix}

\subsection{Datasets}
\label{app: Datasets}

In this section, we describe more details on the datasets used for the experiment.

\noindent \textbf{Molecular Interaction Prediction.}
For the datasets used in the molecular interaction prediction task, we convert the SMILES string into graph structure by using the Github code of CIGIN \cite{pathak2020chemically}.
Moreover, for the datasets that are related to solvation free energies, i.e., MNSol, FreeSolv, CompSol, Abraham, and CombiSolv, we use the SMILES-based datasets provided in the previous work \cite{CombiSolv}. Only solvation free energies at temperatures of 298 K ($\pm$ 2) are considered and ionic liquids and ionic solutes are removed \cite{CombiSolv}.
\begin{itemize}[leftmargin=5mm]
    \item \textbf{Chromophore} \cite{Chromophore} contains 20,236 combinations of 7,016 chromophores and 365 solvents which are given in the SMILES string format. All optical properties are based on scientific publications and unreliable experimental results are excluded after examination of absorption and emission spectra.
    In this dataset, we measure our model performance on predicting \textbf{maximum absorption wavelength (Absorption)}, \textbf{maximum emission wavelength (Emission)} and \textbf{excited state lifetime (Lifetime)} properties which are important parameters for the design of chromophores for specific applications.
    We delete the NaN values to create each dataset which is not reported in the original scientific publications.
    Moreover, for Lifetime data, we use log normalized target value since the target value of the dataset is highly skewed inducing training instability.
    \item \textbf{MNSol} \cite{MNSol} contains 3,037 experimental free energies of solvation or transfer energies of 790 unique solutes and 92 solvents. In this work, we consider 2,275 combinations of 372 unique solutes and 86 solvents following previous work \cite{CombiSolv}.
    \item \textbf{FreeSolv} \cite{FreeSolv} provides 643 experimental and calculated hydration free energy of small molecules in water. In this work, we consider 560 experimental results following previous work \cite{CombiSolv}.
    \item \textbf{CompSol} \cite{CompSol} dataset is proposed to show how solvation energies are influenced by hydrogen-bonding association effects. We consider 3,548 combinations of 442 unique solutes and 259 solvents in the dataset following previous work \cite{CombiSolv}.
    \item \textbf{Abraham} \cite{Abraham} dataset is a collection of data published by the Abraham research group at College London. We consider 6,091 combinations of 1,038 unique solutes and 122 solvents following previous work \cite{CombiSolv}.
    \item \textbf{CombiSolv} \cite{CombiSolv} contains all the data of MNSol, FreeSolv, CompSol, and Abraham, resulting in 10,145 combinations of 1,368 solutes and 291 solvents.
\end{itemize}

\noindent \textbf{Drug-Drug Interaction Prediction.}
For the datasets used in the drug-drug interaction prediction task, we use the positive drug pairs given in MIRACLE Github link\footnote{\url{https://github.com/isjakewong/MIRACLE/tree/main/MIRACLE/datachem}}, which removed the data instances that cannot be converted into graphs from SMILES strings.
Then, we generate negative counterparts by sampling a complement set of positive drug pairs as the negative set for both datasets.
We also follow the graph converting process of MIRACLE \cite{wang2021multi} for the classification task.
\begin{itemize}[leftmargin=5mm]
    \item \textbf{ZhangDDI} \cite{ZhangDDI} contains 548 drugs and 48,548 pairwise interaction data and multiple types of similarity information about these drug pairs.
    \item \textbf{ChChMiner} \cite{ChChMiner} contains 1,322 drugs and 48,514 labeled DDIs, obtained through drug labels and scientific publications.
    \item \textbf{DeepDDI} \cite{DeepDDI} contains 192,284 labeled DDIs and their detailed side-effect information, which is extracted from Drugbank \cite{wishart2018drugbank}.
\end{itemize}

\begin{table}[t]
    \centering
	\small
    \caption{Data statistics. {MI: Molecular Interaction (regression), DDI: Drug-Drug Interaction (classification), SL: Similarity Learning (regression / classification).}}
    \resizebox{0.9\linewidth}{!}{
    \begin{tabular}{cc|cccccc}
    \multicolumn{2}{c|}{Dataset}& $\mathcal{G}^{1}$ & $\mathcal{G}^{2}$ & \# $\mathcal{G}^{1}$ & \# $\mathcal{G}^{2}$ & \# Pairs & Task \\ \hline \hline
    \multicolumn{1}{c|}{Chro-} & Absorption & Chrom. & Solvent & 6416 & 725 & 17276 & MI\\
    \multicolumn{1}{c|}{moph-} & Emission & Chrom. & Solvent & 6412 & 1021 & 18141 & MI \\
    \multicolumn{1}{c|}{ore \tablefootnote{\label{url: Chromophore} \url{https://figshare.com/articles/dataset/DB_for_chromophore/12045567/2}}} & Lifetime & Chrom. & Solvent & 2755 & 247 & 6960 & MI \\ \hline
    \multicolumn{2}{c|}{MNSol \tablefootnote{\url{https://conservancy.umn.edu/bitstream/handle/11299/213300/MNSolDatabase_v2012.zip}}} & Solute & Solvent & 372 & 86 & 2275 & MI \\
    \multicolumn{2}{c|}{FreeSolv \tablefootnote{\url{https://escholarship.org/uc/item/6sd403pz}}} & Solute & Solvent & 560 & 1 & 560 & MI \\
    \multicolumn{2}{c|}{CompSol \tablefootnote{\url{https://aip.scitation.org/doi/suppl/10.1063/1.5000910}}} & Solute & Solvent & 442 & 259 & 3548 & MI \\
    \multicolumn{2}{c|}{Abraham \tablefootnote{\url{https://www.sciencedirect.com/science/article/pii/S0378381210003675}}} & Solute & Solvent & 1038 & 122 & 6091 & MI\\
    \multicolumn{2}{c|}{CombiSolv \tablefootnote{\url{https://ars.els-cdn.com/content/image/1-s2.0-S1385894721008925-mmc2.xlsx}}} & Solute & Solvent & 1495 & 326 & 10145 & MI \\ \hline
    \multicolumn{2}{c|}{ZhangDDI \tablefootnote{\url{https://github.com/zw9977129/drug-drug-interaction/tree/master/dataset}}} & Drug & Drug & 544 & 544 & 40255 & DDI \\
    \multicolumn{2}{c|}{ChChMiner \tablefootnote{\url{http://snap.stanford.edu/biodata/datasets/10001/10001-ChCh-Miner.html}}} & Drug & Drug & 949 & 949 & 21082 & DDI \\
    \multicolumn{2}{c|}{DeepDDI \tablefootnote{\url{https://zenodo.org/record/1205795}}} & Drug & Drug & 1704 & 1704 & 191511 & DDI \\ \hline
    \multicolumn{2}{c|}{AIDS \tablefootnote{\label{url: Similarity} \url{https://github.com/yunshengb/SimGNN}}} & Mole. & Mole. & 700 & 700 & 490K & SL \\
    \multicolumn{2}{c|}{LINUX \textsuperscript{\ref{url: Similarity}}} & Program & Program & 1000 & 1000 & 1M & SL \\
    \multicolumn{2}{c|}{IMDB \textsuperscript{\ref{url: Similarity}}} & Ego-net. & Ego-net. & 1500 & 1500 & 2.25M & SL \\
    \multicolumn{2}{c|}{OpenSSL \tablefootnote{\label{url: Similarity2} \url{https://github.com/runningoat/hgmn_dataset}}} & Flow & Flow & 4308 & 4308 & 18.5M & SL \\ 
    \multicolumn{2}{c|}{FFmpeg \textsuperscript{\ref{url: Similarity2}}} & Flow & Flow & 10824 & 10824 & 117M & SL \\
    \hline
    \end{tabular}}
    \label{tab: data stats}
\end{table}

\noindent \textbf{Graph Similarity Learning.}
For graph similarity learning task, we use three commonly used datasets, i.e., AIDS, LINUX, and IMDB \cite{bai2019simgnn} for regression, and FFmpeg and OpenSSL \cite{xu2017neural} for classification.
\begin{itemize}[leftmargin=5mm]
    \item \textbf{AIDS} \cite{bai2019simgnn} contains 700 antivirus screen chemical compounds and the labels that are related to the similarity information of all pair combinations, i.e., 490K labels. The labels are Graph Edit Distance (GED) scores which are computed with $A^{*}$ algorithm.
    \item \textbf{LINUX} \cite{bai2019simgnn} contains 1,000 Program Dependency Graphs (PDG) generated from the LINUX kernel randomly sampled from the original dataset \cite{wang2012efficient} and the labels that are related to the similarity information of all pair combinations, i.e., 1M labels. The labels are Graph Edit Distance (GED) scores which are computed with $A^{*}$ algorithm.
    \item \textbf{IMDB} \cite{bai2019simgnn} contains 1,500 ego-networks of movie actors/actresses, where there is an edge if the two people appear in the same movie. Labels are related to the similarity information of all pair combinations, i.e., 2.25M labels. The labels are Graph Edit Distance (GED) scores which are computed with $A^{*}$ algorithm.
    \item \textbf{FFmpeg}, \textbf{OpenSSL} \cite{xu2017neural} datasets are generated from popular open-source software FFmpeg \footnote{\url{https://ffmpeg.org/}} and OpenSSL\footnote{\url{https://www.openssl.org/}}, whose graphs denote the binary function's control flow graph. Labels are related to whether two binary functions are compiled from the same source code or not, since the binary functions that are compiled from the same source code are semantically similar to each other.
    In this work, we only consider the graphs that contain more than 50 nodes, i.e., FFmpeg [50, 200] and OpenSSL [50, 200] settings in previous work \cite{zhang2021h2mn}.
\end{itemize}

\subsection{Synthetic Dataset Generation}
\label{App: Synthetic Dataset Generation}
In this section, we describe the detailed process of creating a synthetic dataset.
Following previous work \cite{ying2019gnnexplainer,sui2022causal}, we first generate 16000 synthetic graphs, which are made by combining a causal substructure, which is one of 4 substructures, i.e., "House", "Cycle", "Grid", and "Diamond", and a shortcut substructure, which is one of two substructures, i.e., "Tree" and "BA".
We keep the balance for each combination, i.e., 2000 synthetic graphs for each combination.
Then, we build a paired dataset that consists of a pairwise combination of the synthetic graphs that contain the same shortcut substructure.
Here, we define "positive pair" as a pair that shares the same causal substructure, e.g., \{House, House\} $\rightarrow$ Positive, while "negative pair" is a pair that each graph has a different causal substructure, e.g., \{House, Cycle\} $\rightarrow$ Negative.
Among the generated graph pairs, we randomly sample 10000 positive pairs for each shortcut substructure and 10000 negative pairs for each shortcut substructure, i.e., 40000 pairs in total.
Then, The model is trained to classify whether the pair of graphs is positive or negative within a variety of biasedness.

In this paper, we define the bias as the ratio of positive pairs that contain "BA" substructure among all the positive pairs as follows: 
\begin{equation}
\begin{split}
    \text{bias} (b) &= \frac{\text{Number of positive pairs with BA substructure}}{\text{Number of positive pairs}} \\
    &= \frac{\# \{\text{Causal-BA, Causal-BA}\}}{\#\{\text{Causal-Tree, Causal-Tree}\}+\#\{\text{Causal-BA, Causal-BA}\}}
\end{split}
\end{equation}
where $\# \{\text{Causal-Tree, Causal-Tree}\}$ denotes the number of positive pairs that contain "Tree" shortcut substructure, and $\#\{\text{Causal-BA, Causal-BA}\}$ denotes the number of positive pairs that contain "BA" shortcut substructure.
We set the proportion of negative pairs that contain "BA" substructure to $1-b$.
Therefore, when $b = 0.5$ the dataset is totally unbiased while the bias gets severe as $b$ decreases.
We provide the code for generating synthetic graphs and the paired data and for running the experiment in our anonymized repository.

\subsection{Implementation Details}
\label{App: Implementation Details}
\textbf{Model Training.} In all our experiments, we use the Adam optimizer for model optimization. 
For molecular interaction task and drug-drug interaction task, the learning rate was decreased on plateau by a factor of $10^{-1}$ with the patience of 20 epochs following previous work \cite{pathak2020chemically}.
For similarity learning task, we do not use a learning rate scheduler for the fair comparison with $\text{H}^{2}\text{MN}$ \cite{zhang2021h2mn}.

\noindent \textbf{Hyperparameter Tuning.} 
\label{App: Hyperparameter Tuning}
For fair comparisons, we follow the embedding dimensions and batch sizes of the state-of-the-art baseline for each task.
Detailed hyperparameter specifications are given in Table \ref{tab: Hyperparameters Specifications}.
For the hyperparameters of \proposed, we tune them in certain ranges as follows: learning rate $\eta$ in $\{5e^{-3}, 1e^{-3}, 5e^{-4}, 1e^{-4}\}$ and $\lambda_{1}$ and $\lambda_{2}$ in $\{1e^{-1}, 1e^{-2}, 1e^{-3}, 1e^{-4}, 1e^{-6}, 1e^{-8} \}$.

\begin{table}[h]
    \small
    \centering
    \caption{Hyperparameter specifications ($*$: OOD scenarios).}
    \resizebox{0.9\linewidth}{!}{
        \begin{tabular}{c|c|c|c|c|c|c}
         & Embedding & Batch & \multirow{2}{*}{Epochs} & \multirow{2}{*}{ lr ($\eta$)} & \multirow{2}{*}{$\lambda_{1}$} & \multirow{2}{*}{$\lambda_{2}$} \\
         & Dim ($d$) & Size ($K$) & & & &\\ \hline \hline
         Absorption & 52 & 256 & 500 & 5e-3 & 1e-2 & 1e-3\\
         Emission & 52  & 256 & 500 & 5e-3 & 1e-2 & 1e-2\\
         Lifetime & 52  & 256 & 500 & 1e-3 & 1.0 & 1e-1\\
         MNSol & 42  & 32 & 200 & 1e-3 & 1e-8 & 1e-6\\
         FreeSolv & 42  & 32 & 200 & 1e-3 & 1e-6 & 1e-2 \\
         CompSol & 42  & 256 & 500 & 1e-3 & 1e-2 & 1e-4 \\
         Abraham & 42  & 256 & 500 & 1e-3 & 1e-4 & 1e-8 \\
         CombiSolv & 42  & 256 & 500 & 1e-3 & 1e-4 & 1e-6 \\ \hline
         ZhangDDI & 300  & 512 & 500 & 5e-4 & 1e-3 & 1e-2 \\
         ChChMiner & 300  & 512 & 500 & 5e-4 & 1e-3 & 1e-3 \\ 
         DeepDDI & 300  & 256 & 500 & 5e-4 & 1e-3 & 1e-3 \\\hline
         $\text{ZhangDDI}^{*}$ & 300  & 512 & 500 & 5e-4 & 1e-3 & 1e-1 \\
         $\text{ChChMiner}^{*}$ & 300  & 512 & 500 & 5e-4 & 1e-2 & 1.0 \\ 
         $\text{DeepDDI}^{*}$ & 300  & 256 & 500 & 5e-4 & 1e-2 & 1e-2 \\\hline
         AIDS & 100 & 512 & 10000 & 1e-4 & 1e-2 & 1e-2 \\
         LINUX & 100 & 512 & 10000 & 1e-4 & 1e-2 & 1e-3 \\
         IMDB & 100 & 256 & 10000 & 1e-4 & 1e-3 & 1e-3 \\
         FFmpeg & 100 & 16 & 10000 & 1e-4 & 1e-2 & 1e-2 \\
         OpenSSL & 100 & 16 & 10000 & 1e-4 & 1e-3 & 1e-1 \\ \hline
        \end{tabular}}
    \label{tab: Hyperparameters Specifications}
\end{table}

\subsection{Scaffold Split for Out-of-Distribution}
\label{app: scaffold split for ood}

\begin{figure}[h]
    \centering
    \includegraphics[width=0.8\columnwidth]{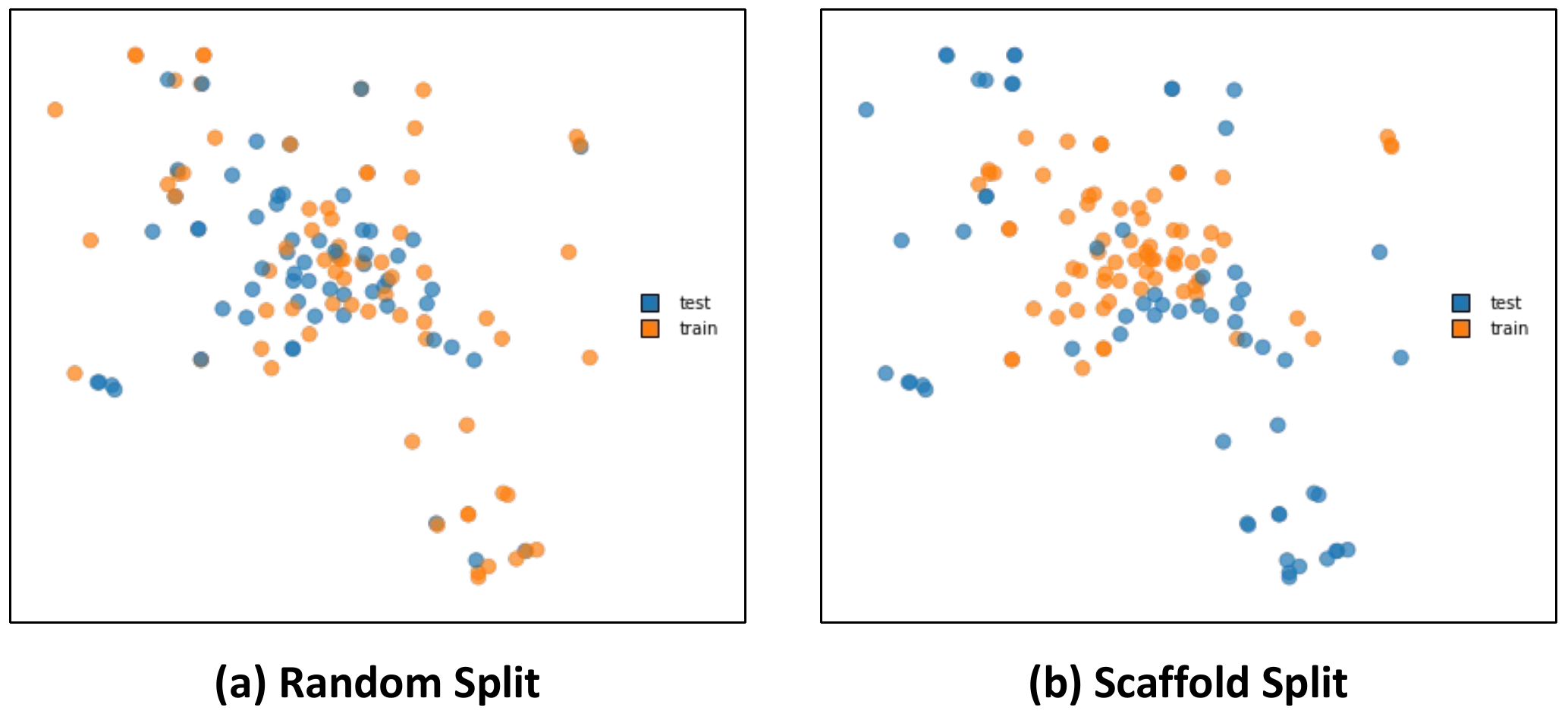} 
    \caption{Comparison of the drug distribution between random split and scaffold split in ZhangDDI dataset.}
    \label{fig: App_data_split}
\end{figure}

In this paper, we verify \proposed's generalization capability on out-of-distribution scenario which is designed by splitting the drugs based on the scaffold.
We argue that molecules belonging to different scaffold classes have totally different distributions as shown in Figure \ref{fig1} (c).
Therefore, as shown in Figure \ref{fig: App_data_split}, we have the similar train and test data distribution when we randomly split the dataset (Figure \ref{fig: App_data_split} (a)), while having different train and test data distribution in scaffold split setting (Figure \ref{fig: App_data_split} (a)), which can also be recognized as out-of-distribution.
For more details on scaffold split, please refer here \footnote{\url{https://oloren.ai/blog/scaff_split.html}}\textsuperscript{,}\footnote{\url{https://www.blopig.com/blog/2021/06/out-of-distribution-generalisation-and-scaffold-splitting-in-molecular-property-prediction/}}.

\end{document}